%% file: main.tex
\documentclass[10pt,twocolumn,letterpaper]{article}

\usepackage{wacv}              


\definecolor{wacvblue}{rgb}{0.21,0.49,0.74}
\usepackage[pagebackref,breaklinks,colorlinks,allcolors=wacvblue]{hyperref}

\usepackage{multirow}
\usepackage{makecell}
\def\wacvPaperID{451} 
\def\confName{WACV}
\def\confYear{2027}

\title{Implicit Neural Representation-Based Continuous Single Image Super-Resolution: An Empirical Benchmark}

\author{Tayyab Nasir\\
The University of Western Australia\\
Perth, Western Australia\\
{\tt\small tayyabnasir22@gmail.com, tayyab.nasir@research.uwa.edu.au}
\and
Daochang Liu\\
The University of Western Australia\\
Perth, Western Australia\\
{\tt\small daochang.liu@uwa.edu.au}
\and
Ajmal Mian\\
The University of Western Australia\\
Perth, Western Australia\\
{\tt\small ajmal.mian@uwa.edu.au}
}

\begin{document}
\maketitle
\input{sec/0_abstract}

\input{sec/1_introduction}

\input{sec/2_background}

\input{sec/3_method}

\input{sec/4_results}

\input{sec/5_conclusion}

\appendix
\input{sec/X_suppl}

\clearpage
{
    \small
    \bibliographystyle{ieeenat_fullname}
    \bibliography{main}
}

\end{document}

%% file: sec/0_abstract.tex
\begin{abstract}
Implicit neural representation (INR) has become the standard approach for arbitrary-scale image super-resolution (ASSR). However, no systematic empirical study has examined the effectiveness of existing methods under consistent conditions, nor investigated the effects of different training recipes, such as objective design, optimization strategies, and scaling behavior. A rigorous empirical analysis is essential not only for benchmarking performance and revealing true gains but also for establishing the current state of ASSR, identifying saturation limits, and highlighting promising directions. We fill this gap by training 6 INR-based ASSR methods under 9 controlled training recipes and evaluating them across 7 datasets and 7 image quality assessment (IQA) metrics, presenting aggregated performance results through a unified ranking framework that enables more reliable interpretation of performance comparisons and evaluation claims. To facilitate reproducible comparisons, a unified codebase is also provided\footnote{Code available at: \href{https://github.com/tayyabnasir22/INR-ASSR-Emperical-Analysis}{https://github.com/tayyabnasir22/INR-ASSR-Emperical-Analysis}}. Furthermore, we investigate the impact of carefully controlled training configurations on perceptual image quality and analyze the role of auxiliary objectives in preserving edges, textures, and fine details during training. Our analysis yields the following insights, previously overlooked: (1) Recent, more complex INR methods provide only marginal improvements over earlier methods, indicating architectural saturation on existing benchmarks. (2) Model performance is strongly correlated with training configurations, a factor neglected in prior comparisons. (3) Auxiliary objectives consistently enhance texture fidelity across architectures compared to standard L1-Loss, emphasizing the role of objective design for targeted perceptual gains. (4) INR-based ASSR exhibits consistent, monotonic scaling behavior across model capacity, training compute, and data diversity, though with diminishing returns as complexity grows.
\end{abstract}

%% file: sec/1_introduction.tex
\section{Introduction}
\label{sec:intro}
Single-image super-resolution (SISR) is the process of converting a low-resolution image into a higher-resolution one~\cite{moser2023hitchhiker, niu2020single, moser2024diffusion, wang2020deep}. In its simplest form, such a process deals only with creating a fixed-scale version of the image, usually upscaling by an integer scale. Arbitrary scale, or continuous, super-resolution can generate a higher-resolution image of any size, using non-integer scaling factors~\cite{liu2024arbitrary, chen2021learning, wei2023super}. Implicit neural representation (INR) methods have gained significant attention for continuous image super-resolution, as their ability to map continuous signal coordinates directly to RGB values of higher resolution provides a simple and flexible way of training~\cite{mildenhall2021nerf, chen2019learning, atzmon2020sal}. This paper focuses on continuous image super-resolution using INR-based methods, emphasizing empirical analysis over methodological novelty. We present a systematic evaluation framework and provide insights that are valuable for future research. We focus only on supervised INR-based ASSR techniques to ensure a consistent evaluation setting, avoiding confounding factors from unsupervised, semi-supervised, and zero-shot approaches. 

MetaSR~\cite{hu2019meta} and LIIF~\cite{chen2021learning} are two of the earliest methods that used INR-based techniques for continuous single-image super-resolution, where the former uses it as an upscaling module, while the latter defines an encoder-decoder INR architecture. The encoder generates a latent feature map in spatial dimensions from the input low-resolution image. Conventionally, fixed SISR models are used as an encoder~\cite{lim2017enhanced, zhang2018residual}. The generated latent code is queried using the desired higher-resolution coordinates. The resultant latent feature vectors, along with additional scale information, are passed to the decoder to get the RGB values for each of the input coordinates.  Additional components are also used within the same architecture~\cite{chen2021learning}, such as feature unfolding to enhance the quality of information encoded in the latent representations, cell decoding to represent the target scale information, and local ensembling for averaging over neighboring coordinates.

Techniques like LTE~\cite{lee2022local}, SRNO~\cite{wei2023super}, CLIT~\cite{chen2023cascaded}, CiaoSR~\cite{cao2023ciaosr}, and HIIF~\cite{jiang2025hiif} follow the aforementioned architecture. Each technique, as reported in prior research, compares its performance with its predecessors in a semi-controlled experimental setting, using the same datasets and scaling factors, with Peak Signal-to-Noise Ratio (PSNR) as the primary evaluation metric. However, a lack of transparency in the reproducibility of results and fair benchmarking remains problematic, as newer studies do not replicate prior results under comparable settings. For example, some studies use a larger patch size of $128^{2}$ and the Stochastic Gradient Descent with Warm Restarts (SGDR)~\cite{loshchilov2016sgdr} for training, rather than the $48^{2}$ patch and multi-step training scheduler used in prior studies. Hence, it remains unclear whether the reported improvements reflect true architectural improvements or are largely due to differences in training strategies, and whether the earlier techniques, being simpler, can perform well compared to the newer, more complex works, under similar settings. 

Another limitation lies in the lack of diverse image quality assessment (IQA), with most studies relying solely on PSNR, thereby offering a limited perspective on perceptual image quality. Validation PSNR is often used to select the best model weights in each study, introducing further bias in comparative evaluations. Moreover, the impact of different training recipes on targeted perceptual improvements for INR-based ASSR remains unexplored. These issues are largely overlooked in prior works and continue to pose a significant bottleneck to further advancement in this area.

To address the lack of consistency in training and evaluation of existing techniques, and to introduce general transparency in comparisons, we conduct extensive experiments using diverse but consistent settings for training and evaluation. Specifically, we trained 6 ASSR methods using 9 experimental setups and 2 different encoders, and evaluated them across 7 datasets and 7 IQA metrics. Furthermore, we provide a centralized code repository for easy reproducibility and comparison. Additionally, we investigate the impact of different learning strategies on performance by employing different training recipes and objectives, demonstrating the potential for achieving targeted perceptual quality improvements under carefully designed training settings. Finally, we present a comprehensive set of analyses, addressing the aforementioned questions. A summary of our key findings is as follows: 

\begin{itemize}
\item \textbf{Marginal Improvements of Recent Models:} Recent, more complex INR-based ASSR models yield marginal improvements over earlier methods, suggesting that the current architectural designs may have reached a saturation point on existing datasets. The maximum PSNR gain observed between the best model (SRNO) and the runner-up was merely 0.035 dB, further supporting this observation.

\item \textbf{Sensitivity to Training Configurations:} Performance gains can be attained not only through architectural modifications but also by carefully tuning the training configurations and optimization settings. For example, HIIF shows a strong dependency on the learning rate scheduler. Hence, reproducing prior work under new training settings is essential for a fair comparison.

\item \textbf{Sensitivity to Perceptual Objectives:} Different architectures show varying performance trends across IQA metrics. Moreover, training objectives targeting specific perceptual attributes present a promising direction for further progress in INR-based super-resolution. This is supported by the improved texture fidelity achieved with a simple auxiliary gradient-consistency term, adopted from prior work as a controlled instrument, over the standard L1 loss.

\item \textbf{Scaling Behavior and Diminishing Returns:} INR-based ASSR exhibits consistent, monotonic scaling behavior across model capacity, training compute, and data diversity, with performance improving reliably along each axis. However, the rate of improvement diminishes as complexity grows, indicating a saturation point in parameter-driven gains, consistent with the architectural saturation observed elsewhere in our analysis.
\end{itemize}
In summary, we provide a systematic empirical analysis that consolidates existing efforts and provides a foundation for future studies on ASSR. Our main contributions include:
\begin{enumerate}
\item We present a systematic audit of existing INR-based continuous super-resolution methods, examining the effects of different training strategies, including variations in optimization schemes and objective designs.

\item We present a unified benchmarking framework that enables refined evaluation and transparent interpretation of prior claims. Rigorous stress testing across a comprehensive set of IQA metrics reveals performance saturation in INR-based ASSR on the DIV2K benchmark.

\item We create a single code repository for a centralized, systematic benchmarking framework to ensure fair evaluations and facilitate research extensions.

\end{enumerate}

%% file: sec/2_background.tex
\section{Related work}
\label{sec:related}
\textbf{Single-Image Super-Resolution (SISR):} SRCNN~\cite{dong2014learning} was the first method that used a deep neural network architecture for SISR, proposing a fully convolutional neural network. SRCNN inspired a series of shallow and deep CNN-based architectures for SISR~\cite{kim2016accurate, zhang2017beyond, zhang2017learning, dong2016accelerating, shi2016real}. Recently, attention~\cite{dai2019second, niu2020single}, diffusion~\cite{li2022srdiff, saharia2022image}, and state space model (SSM)~\cite{guo2024mambair, ren2024mambacsr, shi2025vmambair} based techniques have also been proposed for SISR. However, many of these architectures rely on components such as deconvolution~\cite{dong2016accelerating} or sub-pixel convolutions~\cite{shi2016real}, which constrain the upsampling process to fixed integer scales, limiting their flexibility for continuous-scale image reconstruction~\cite{yao2023local, chen2023cascaded}. 

\textbf{Arbitrary-Scale Super-Resolution (ASSR)} extends conventional SISR by enabling upscaling by any non-integer factor, making it a more practical and flexible solution for real-world applications. VDSR~\cite{kim2016accurate} can be viewed as an early step toward ASSR, where its pre-upsampling strategy resizes the input to the target scale via bicubic interpolation, after which a CNN reconstructs high-frequency details, decoupling the network from a fixed upscaling factor. Subsequent attempts, such as OverNet~\cite{behjati2021overnet}, EQSR~\cite{wang2023deep}, and SRWarp~\cite{son2021srwarp}, further explore pre-upsampling and warping-based strategies. A detailed discussion of these approaches is presented in the survey~\cite{liu2024arbitrary}.

\textbf{Implicit Neural Representation (INR):} Over the years, INR has emerged as the most popular approach for ASSR, owing to its simplicity and flexibility. INR utilizes a multi-layer perceptron (MLP) for mapping input coordinates to a continuous signal~\cite{mildenhall2021nerf}. It has been widely used for 3D tasks, such as scene reconstruction, shape modeling, and compression~\cite{genova2019learning, chen2019learning, mescheder2019occupancy}, as well as continuous image representation~\cite{klocek2019hypernetwork}. MetaSR~\cite{hu2019meta} was the first to use coordinate and scale information for ASSR, introducing a meta-upscale module that dynamically generates convolutional weights to produce high-resolution images. 
LIIF~\cite{chen2021learning} introduced an MLP-based local implicit function to predict RGB values for queried coordinates, proposing the first encoder-decoder architecture for INR-based ASSR, serving as the basis for later methods, including LTE~\cite{lee2022local}, SRNO~\cite{wei2023super}, CiaoSR~\cite{cao2023ciaosr}, CLIT~\cite{chen2023cascaded}, and HIIF~\cite{jiang2025hiif}. Additional architectural components, such as feature unfolding and local ensemble mechanisms, are also used in INR-based architectures for ASSR~\cite{chen2021learning}.
LTE proposed changes for improving fine detail reconstruction by transforming coordinate features into the Fourier domain before feeding them into the implicit decoder. CiaoSR and CLIT introduced the use of attention for context-aware feature encoding for the query coordinates, while SRNO uses Galerkin attention to capture non-local spatial dependencies more efficiently. HIIF introduced hierarchical positional encoding to represent local features and coordinates across multiple scales.

\textbf{Training and Evaluation Objectives:} Different objectives have been applied for learning and evaluating super-resolution techniques. L1-loss is the most commonly used loss for ASSR, particularly in INR-based methods, as summarized in Table~\ref{tab:prevconfigs}. However, it has a limited ability to reconstruct high-frequency details. Alternate objective functions targeting perceptual reconstruction have also been explored. Perceptual losses~\cite{johnson2016perceptual} compute the Euclidean distance between high-level features extracted from a pretrained network for the ground truth and generated images, encouraging perceptual similarity. In this category, texture losses, often based on the Gram matrix~\cite{gatys2015neural}, promote consistency in texture and structural details. Similarly, mixed losses combine L1-loss with perceptual or texture loss, aiming to balance pixel-level accuracy with perceptual fidelity.

PSNR is the most widely used metric for evaluating super-resolution methods. While it strictly measures pixel-level intensity differences, it does not capture perceptual similarity. More perception-oriented metrics include SSIM~\cite{wang2004image}, which evaluates luminance, contrast, and structural similarity, and LPIPS~\cite{zhang2018unreasonable}, which leverages deep neural network features to quantify perceptual similarity between image patches. LPIPS generally aligns more closely with human perception of image quality than PSNR. Nevertheless, despite its limitations, PSNR remains the primary evaluation metric, particularly for INR-based ASSR methods, limiting the assessment of perceptual performance.

\textbf{Empirical Studies and Benchmarks:} Systematic empirical analyses have proven valuable in adjacent areas, including reproducible scaling studies for contrastive language-image learning~\cite{cherti2023reproducible} and scene text recognition~\cite{rang2024empirical}, as well as community benchmarks such as the NTIRE challenges for super-resolution~\cite{agustsson2017ntire, li2023ntire}. However, no comparable study exists for INR-based ASSR, in which methods continue to be compared across heterogeneous training recipes and a narrow set of IQA metrics.

%% file: sec/3_method.tex
\section{Method}
\label{sec:method}
Our proposed unified experimental framework systematically and consistently evaluates INR-based ASSR methods across diverse settings. It provides a consistent performance-ranking protocol across a comprehensive set of evaluation metrics and datasets, presents a targeted analysis of perceptual improvement gains from tuned objectives, and examines scaling behavior across model capacity, training compute, and data diversity.

\subsection{Datasets}
Following prior works, we used the DIV2K~\cite{agustsson2017ntire} dataset, which consists of 800 high-quality 2K-resolution images for training and 100 images for validation. In the context of INR-based ASSR, the effective sample size is determined by the number of sampled pixel coordinates per image, where each high-resolution image can be randomly sampled for a large number of pixel-coordinate pairs, providing substantial diversity for optimizing deeper networks. For out-of-distribution testing, we used Set5~\cite{bevilacqua2012low}, Set14~\cite{zeyde2010single}, BSD100~\cite{martin2001database}, and Urban100~\cite{huang2015single} datasets, along with SVT~\cite{wang2010word} and CelebA-HQ~\cite{karras2017progressive} for diverse domain-specific evaluations given in the Appendix. The combined test sets are denoted as $D$.

\subsection{Empirical Experiment Settings}
\textbf{Methods and Configurations:} We conducted comprehensive experiments on existing INR-based methods, including MetaSR, LIIF, LTE, SRNO, CiaoSR, and HIIF. This set of techniques constitutes the model set $M$. Another related work, CLIT~\cite{chen2023cascaded}, was excluded because its multi-stage cascaded training strategy is incompatible with our single-stage unified training protocol, and its parameter count is roughly$15 \times$ that of LIIF, as including it would confound architectural comparison with training-procedure and capacity differences. Its reported performance is nevertheless comparable to CiaoSR, SRNO, and HIIF~\cite{cao2023ciaosr, wei2023super, jiang2025hiif}.

\textbf{Training Details:} All models were trained for 150 epochs to keep the analysis computationally manageable. The reduced training duration does not compromise convergence, as detailed convergence plots across different models and training settings are provided in the Appendix, confirming that all models reached stability well before 150 epochs. Moreover, our results closely match those reported in prior work using 1000 training epochs. For example, LTE at $12 \times$ scaling on DIV2K is reported as 23.78 dB with 1000 epochs, compared to 23.60 dB in our setting. This residual gap from extended training is expected to affect all methods similarly, since all models are trained under identical budgets. Our conclusions rest on relative comparisons under matched conditions rather than absolute parity with prior results.

Table~\ref{tab:prevconfigs} presents a comparison between the training and evaluation settings adopted in prior work and those used in our unified framework. Our framework has a broader and more diverse set of training recipes, denoted as $T$, along with comprehensive evaluation criteria.

\begin{table*}[!htb]
  \centering
  \footnotesize
  \begin{tabular}{ccccccccc}
\toprule
Recipe Name & Model & Loss & Patch Size & Scale Range & LR Scheduler & Batch & Optimizer & Reported IQA \\
\midrule
& MetaSR & L1-Loss & 50 & 1–4 & Multi-Step & 16 & Adam & PSNR, SSIM \\
& LIIF   & L1-Loss & 48 & 1–4 & Multi-Step & 16 & Adam & PSNR \\
& LTE    & L1-Loss & 48 & 1–4 & Multi-Step & 16 & Adam & PSNR \\
& SRNO   & L1-Loss & 128 & 1–4 & SGDR & 64 & Adam & PSNR \\
& CiaoSR & L1-Loss & 48 & 1–4 & Multi-Step & 16 & Adam & \makecell{PSNR, SSIM, \\ LPIPS} \\
& HIIF   & L1-Loss & 48 & 1–4 & SGDR & 16 & Adam & PSNR \\
\midrule
L1-Loss & \multirow{9}{*}{\makecell{
    MetaSR, \\ LIIF, \\ LTE, \\ SRNO, \\ CiaoSR, \\ HIIF 
 }}& L1-Loss & 48 & 1–4 & Multi-Step & \multirow{9}{*}{\makecell{32 for \\ EDSR, \\ 16 for \\ RDN}} & Adam &  \multirow{9}{*}{\makecell{PSNR, \\ SSIM, \\ GMSD, \\ FSIM, \\ VIF, \\ SR-SIM, \\ LPIPS}} \\
SGDR & & L1-Loss & 48 & 1–4 & SGDR & & Adam  &  \\
AdamW & & L1-Loss & 48 & 1–4 & Multi-Step & & AdamW  &  \\
Large Patch & & L1-Loss & 64 & 1–4 & Multi-Step & & Adam &  \\
Larger Patch & & L1-Loss & 128 & 1–4 & Multi-Step & & Adam &  \\
Larger Scale &  & L1-Loss & 64 & 1–6 & Multi-Step & & Adam  & \\
Gradient Loss & & Gradient Loss & 48 & 1–4 & Multi-Step & & Adam & \\
Gram-L1-Loss & & Gram-L1 Loss & 48 & 1–4 & Multi-Step & & Adam & \\
VGG-Loss & & VGG Loss & 48 & 1–4 & Multi-Step & & Adam & \\
\bottomrule
  \end{tabular}
  \caption{Comparison of training configurations between prior ASSR methods (top) and our unified experimental framework (bottom). The last nine rows detail our standardized training recipes $T$, which combine different loss designs, scale ranges, and schedulers for comprehensive evaluation across multiple IQA metrics.}
  \label{tab:prevconfigs}
\end{table*}

\textbf{Experimental Coverage and Evaluations:} We conducted a total of 108 experiments, covering 6 techniques, 2 encoder variants~\cite{lim2017enhanced, zhang2018residual}, and 9 training recipes. Additionally, 2 supplementary training configurations were used to analyze the effects of batch size variations for the training compute analysis. Validation PSNR was tracked during training, and both the best and the 150-epoch model weights were retained for aggregate performance analysis. Performance of each technique was evaluated across multiple datasets and scales using seven IQA metrics, including PSNR, SSIM~\cite{wang2004image}, GMSD~\cite{xue2013gradient}, FSIM~\cite{zhang2011fsim}, VIF~\cite{sheikh2005visual}, SR-SIM~\cite{zhang2012sr}, and LPIPS~\cite{zhang2018unreasonable}, which together define the evaluation set $Q$. Evaluation is performed across scaling factors $s \in S = \{2, 3, 4, 6, 12, 18, 24, 30\}$, with dataset-specific subsets detailed in the Appendix.

\subsection{Aggregated Ranking Framework}
\label{subsec:bordaframework}
We adopt a ranking scheme based on Borda count~\cite{mclean1990borda} to provide an aggregated view of each model’s performance across different scales, datasets, IQA metrics, and training recipes. For ranking at the level of individual training recipes, let $V_{m}^{d,s,q}$ denote the value-rank of model $m \in M$ evaluated on dataset $d \in D$, scale $s \in S$, and IQA metric $q \in Q$, where a higher value-rank indicates better performance. The Borda count for model $m$ is defined as:

{\small
\begin{equation}
B_{m} = \sum_{d}^{D}\sum_{s}^{S}\sum_{q}^{Q} V_{m}^{d,s,q}. 
\end{equation}
}

The corresponding aggregated ranking $r_{m}$ for each model is then computed as:

{\small
\begin{equation}
r_m = \operatorname*{rank}_{\text{desc}}(B_m).
\end{equation}
}

\noindent where $r_{m} \in \mathbb{N}$ denotes the rank of model $m$, and $R_{M} = \left\{ r_{m} | m \in M \right\}$ represents the complete ranking of all models, with a lower value indicating superior overall performance. Similarly, to obtain rankings based on individual IQA metrics, we define $V_{m}^{d,s,t}$, where $t \in T$ represents the training recipe, and apply the same Borda count aggregation procedure to compute the ranks. Finally, we use $V_{m}^{d,s,q,t}$ with the same aggregation procedure, incorporating all datasets, scales, metrics, and training recipes, to obtain the comprehensive global model ranking. The proposed framework evaluates existing techniques and evaluation practices by aggregating performance across multiple IQA metrics and analyzing their sensitivity and robustness under varying training conditions.

%% file: sec/4_results.tex
\section{Results and Analysis}
\label{sec:results}

\subsection{Marginal Improvements}
\begin{figure}[!htb]
  \centering
   \includegraphics[width=\linewidth,]{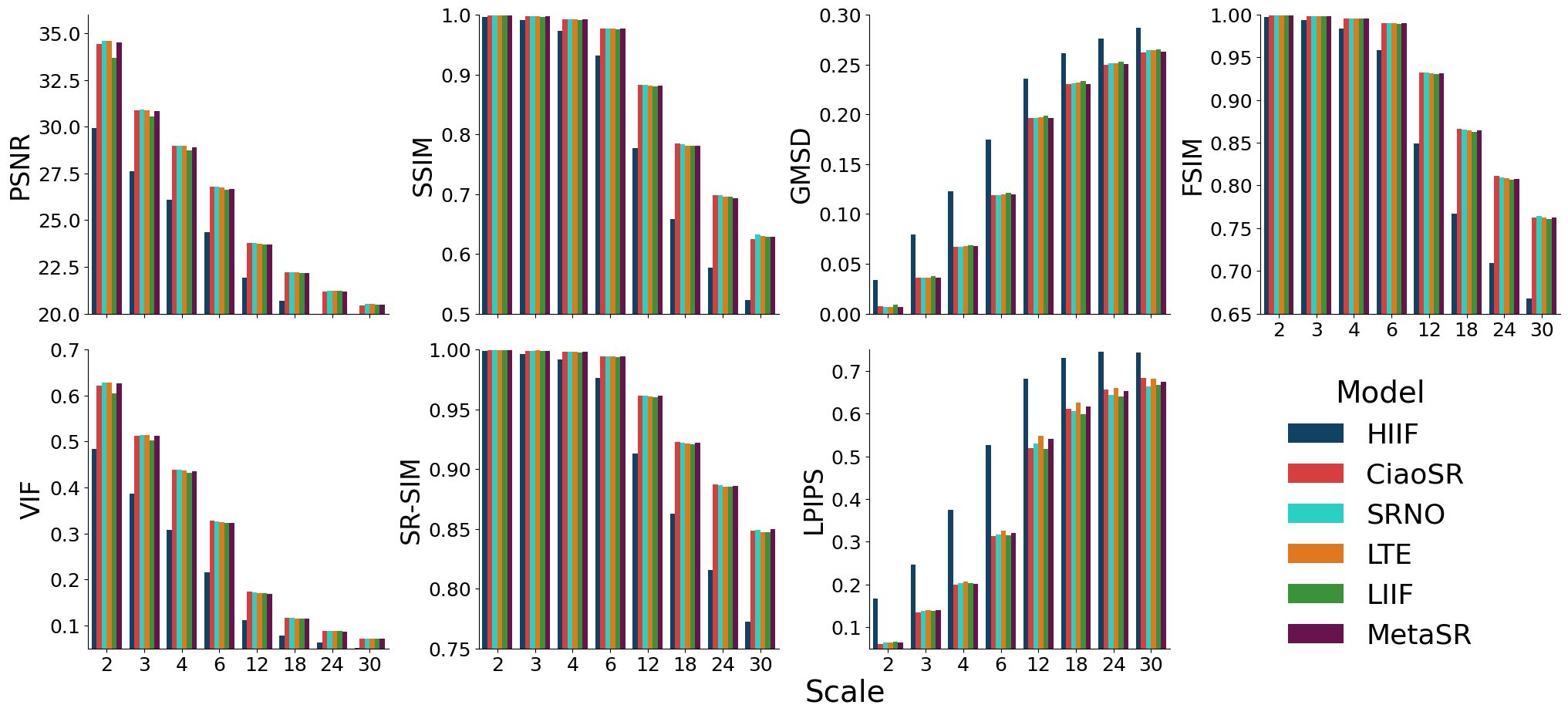}
   \caption{Best results achieved by each model on the DIV2K validation set across scaling factors (x-axis), with one subplot per IQA metric (y-axis: metric value). Only marginal performance differences are observed across architectures at all scales. Note that y-axis ranges are deliberately kept narrow to make sub-saturation differences visible.}
\label{fig:marginal}
\end{figure}

Figure~\ref{fig:marginal} compares the best results achieved by each model across all evaluation metrics on the DIV2K validation set, using the EDSR encoder. For this comparison, we selected each model’s best-performing trained instance, so that every architecture is represented by its most favorable configuration. The results indicate that while recent methods have steadily progressed and introduced valuable architectural refinements, the overall performance gains across scales are marginal. This suggests that INR-based ASSR architectures have matured considerably on existing benchmarks, highlighting the need for more disruptive architectural advances and more diverse and challenging datasets. To quantify the degree of saturation, we compute mean performance and 95\% confidence intervals (CI) for each method, across 3 random training seeds and across training recipes (full tables in the Appendix). At $12 \times$ on DIV2K, seed-level CIs are at most $\pm 0.005$ dB PSNR, with the top three methods separated by only 0.01 dB (23.62--23.63 dB). Across training recipes, the top three remain within 0.05 dB of one another. The sole exception is HIIF, whose wider recipe-level intervals reflect a strong scheduler sensitivity analyzed in Section~\ref{subsec:sensitivity}. Together with Figure~\ref{fig:marginal}, these results indicate that differences between architectures are small compared to those induced by training configuration, and that incremental architectural refinements are yielding diminishing improvements.

\subsection{Sensitivity to Training Configurations}
\label{subsec:sensitivity}
Figure\hyperref[fig:ranked]{~\ref*{fig:ranked}a)} presents an aggregated analysis, summarizing the consistency, stability, and robustness of different techniques under diverse training configurations, computed separately for each encoder via the Borda framework~\ref{subsec:bordaframework} using the value-rank $V_{m}^{d,s,q}$. The resulting global rank for each technique is shown at the right end of the figure, with additional analysis presented in the Appendix. The fact that different models excel under different configurations highlights the sensitivity of specific architectures to particular training strategies, emphasizing the need for fair and consistent experimental comparisons. The key takeaway is that even earlier, simpler models can outperform more recent approaches when trained under appropriately tuned settings. To further investigate this behavior, we provide a detailed analysis of the following three specific training factors:

\begin{figure}[!htb]
  \centering
   \includegraphics[width=1\linewidth]{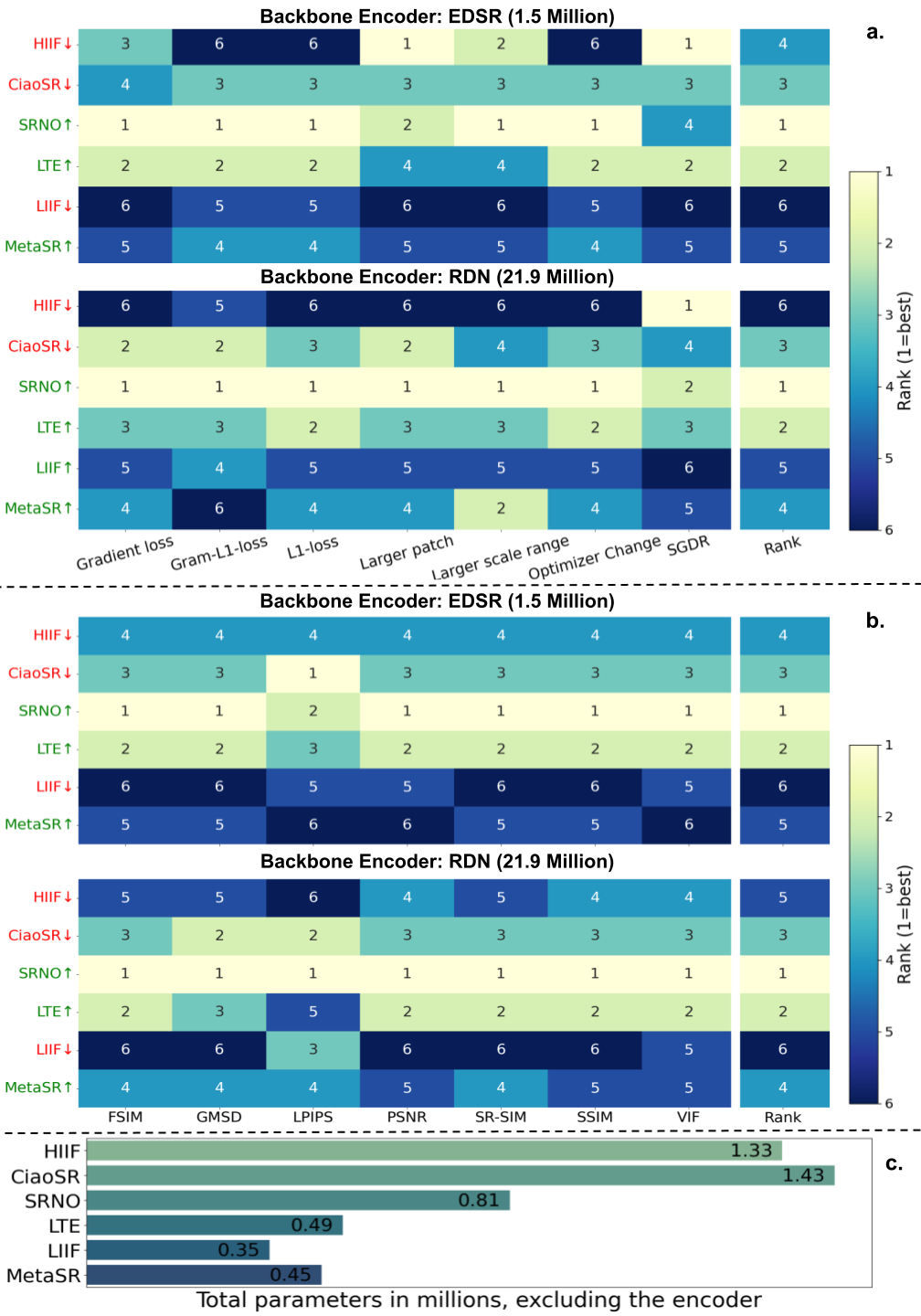}
   
   \caption{\textbf{a)} Models ranked based on aggregated performance across datasets, scales, and seven IQA metrics using Borda count aggregation, where lower ranks indicate better overall performance. Variations in rankings highlight the sensitivity of different techniques to training configurations. The final rank for each method is shown on the right, and red downward arrows indicate a drop in rank compared to previously reported results~\cite{jiang2025hiif}. 
   \textbf{b)} Models ranked per IQA metric based on aggregated performance across datasets, scales, and settings using Borda count aggregation. Variations in ranking highlight the sensitivity of different techniques to specific perceptual metrics.
   \textbf{c)} Additionally, we report the total number of parameters for each technique, with a more detailed model complexity analysis provided in the Appendix.}
   \label{fig:ranked}
\end{figure}

\begin{figure*}[!htb]
    \centering
    \includegraphics[width=\linewidth]{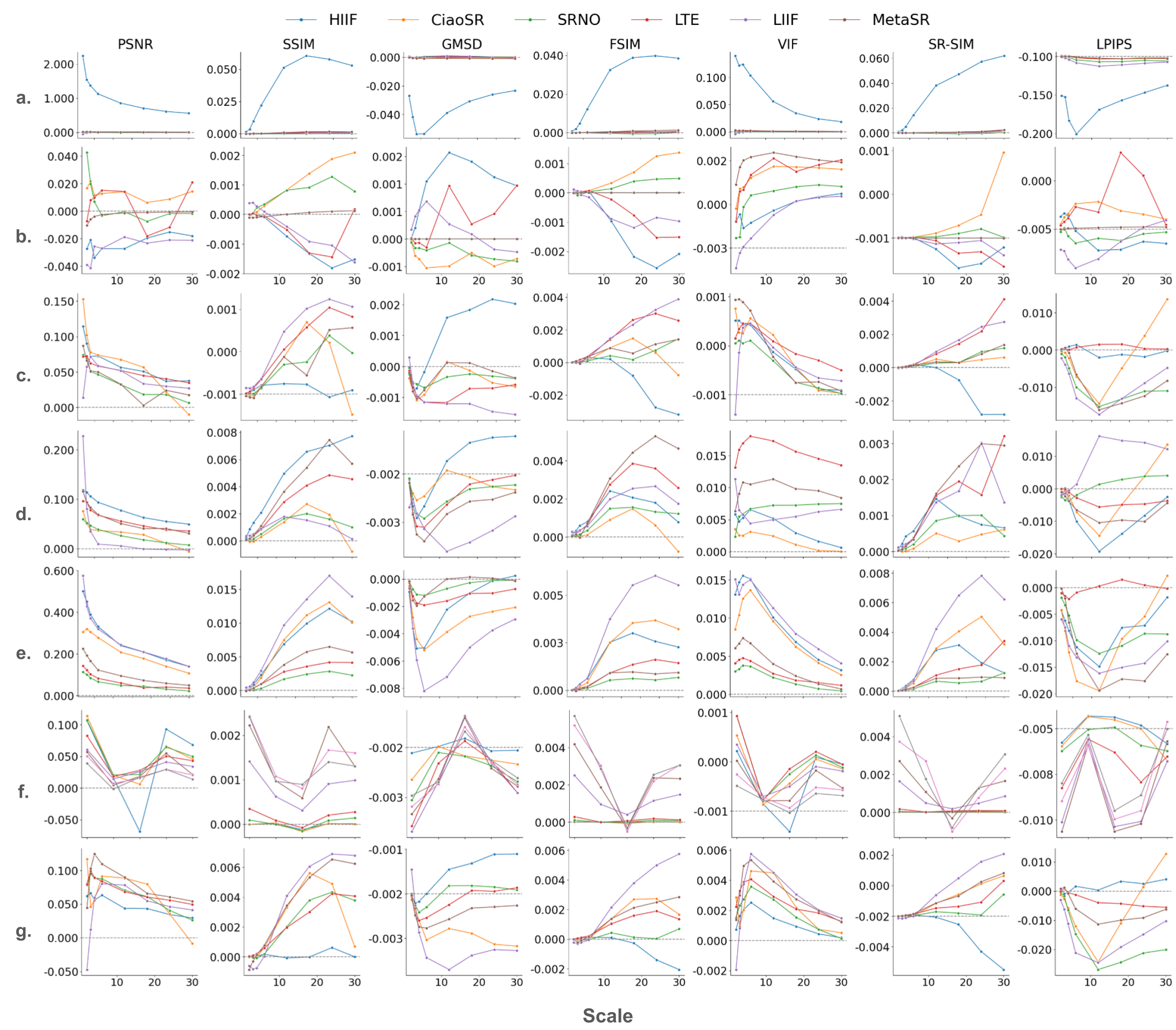}
\caption{Change in each IQA metric (vertical axis) induced by a change in training configuration, plotted against scale. For PSNR, SSIM, FSIM, VIF, and SR-SIM, higher is better, while for GMSD and LPIPS, lower is better. \textbf{a)} Switching from a multi-step scheduler to SGDR~\cite{loshchilov2016sgdr} yields a general improvement across metrics and techniques, with HIIF showing the largest gain. \textbf{b)} Replacing Adam with AdamW~\cite{loshchilov2017decoupled} benefits CiaoSR notably but does not generalize to other techniques. \textbf{c, d)} Increasing the training patch size from $48^{2}$ to $64^{2}$ (c) and $128^{2}$ (d) yields a general improvement across metrics. \textbf{e)} Switching the encoder from EDSR (1.5M parameters) to RDN (21.9M) improves performance, consistent with capacity scaling. \textbf{f)} Extending training from 100 to 150 epochs improves performance, consistent with compute scaling. \textbf{g)} Widening the training scale range from 1--4 to 1--6 improves performance, consistent with data-diversity scaling.}
    \label{fig:iqa_results}
\end{figure*}

\textbf{Impact of Learning Rate Scheduler:} Figure\hyperref[fig:iqa_results]{~\ref*{fig:iqa_results}a)} presents a delta plot reporting the change in performance for each model when switching from a multi-step learning rate scheduler, which decays the learning rate at fixed intervals, to an SGDR~\cite{loshchilov2016sgdr}. Although all the techniques show a general improvement under the SGDR, HIIF shows the most significant gain. This aligns with prior reports~\cite{jiang2025hiif}, where HIIF’s superior performance was achieved using SGDR. However, the fact that other models also benefit from this updated scheduler reinforces the importance of conducting fair comparisons under fixed training settings.

\textbf{Impact of Optimizer:} Figure~\hyperref[fig:iqa_results]{\ref*{fig:iqa_results}b)} presents a delta plot showing the change in performance when replacing the Adam optimizer, used in prior work across all six techniques, with AdamW~\cite{loshchilov2017decoupled}. The results indicate a negative impact for most techniques, with improvements observed only for CiaoSR and, for certain IQA metrics, SRNO. This suggests that the effect of optimizer choice is architecture-dependent rather than uniform, underscoring its importance for reproducibility and fair comparison.

\textbf{Impact of Input Patch Size:} Figure\hyperref[fig:iqa_results]{~\ref*{fig:iqa_results}c)} shows a delta plot reporting the change in performance when switching from an input patch size of $48^{2}$ to $64^{2}$. Similarly, Fig.\hyperref[fig:iqa_results]{~\ref*{fig:iqa_results}d)} plots the change in performance moving to an input patch size of $128^{2}$. There is a consistent trend of improvement with larger patch sizes across architectures. We regard this as a general effect of the enlarged spatial sampling rather than an architecture-specific advantage, as no single method benefits disproportionately.

\subsection{Sensitivity to Perceptual Objectives}
\textbf{Impact of IQA Metrics on Model Rankings:} Figure\hyperref[fig:ranked]{~\ref*{fig:ranked}b)} provides an aggregated overview of each technique's performance per IQA metric, computed with the Borda framework using the value-rank $V_{m}^{d,s,t}$. Rankings vary considerably across metrics, indicating that each architecture is sensitive to different perceptual characteristics. As summarized in Table~\ref{tab:prevconfigs}, most prior works have evaluated performance using PSNR alone, which limits the understanding of the perceptual trade-offs. Incorporating a wider range of IQA metrics enables a more balanced and comprehensive evaluation, providing deeper insight into how different architectures optimize structural accuracy and perceptual realism.

\begin{figure}[!htb]
  \centering
   \includegraphics[width=\linewidth]{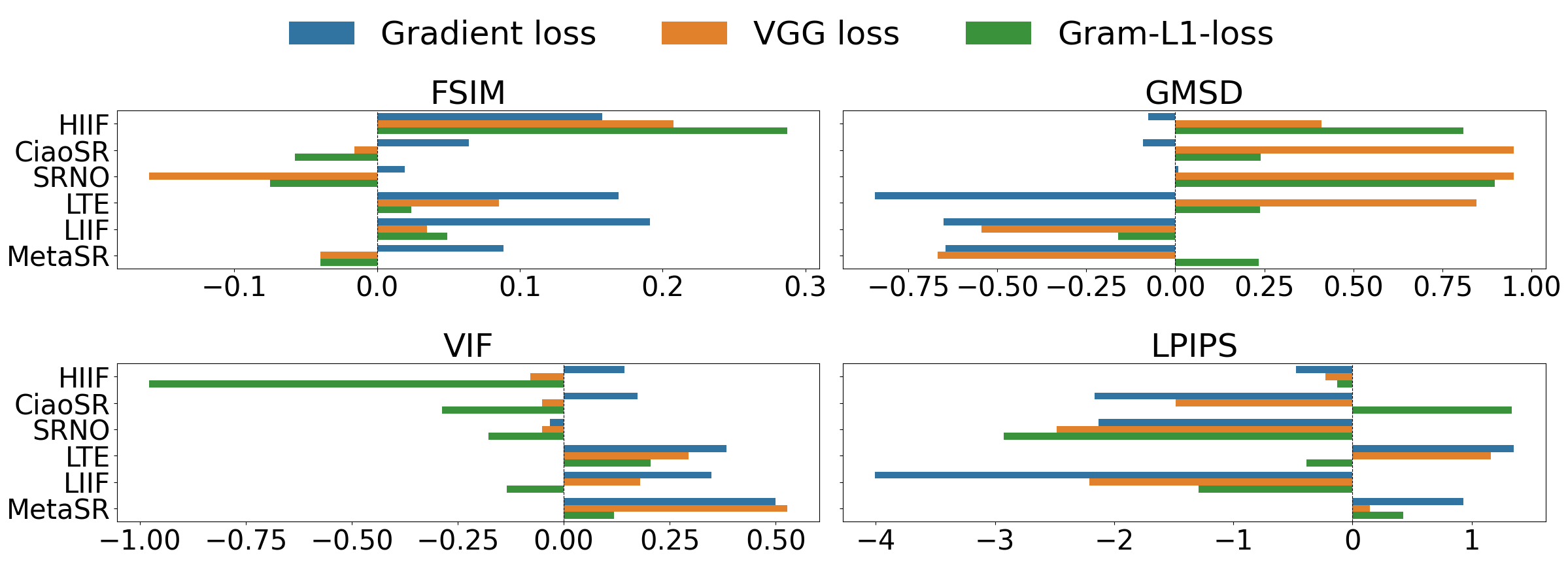}

   \caption{Relative change in texture-sensitive IQA metrics when replacing the L1 loss with L1-Gram, VGG, and Hybrid-Gradient losses. Overall, the Hybrid-Gradient loss shows consistent performance improvements across architectures.}
   \label{fig:losschange}
\end{figure}

\begin{figure}[!htb]
  \centering
   \includegraphics[width=\linewidth]{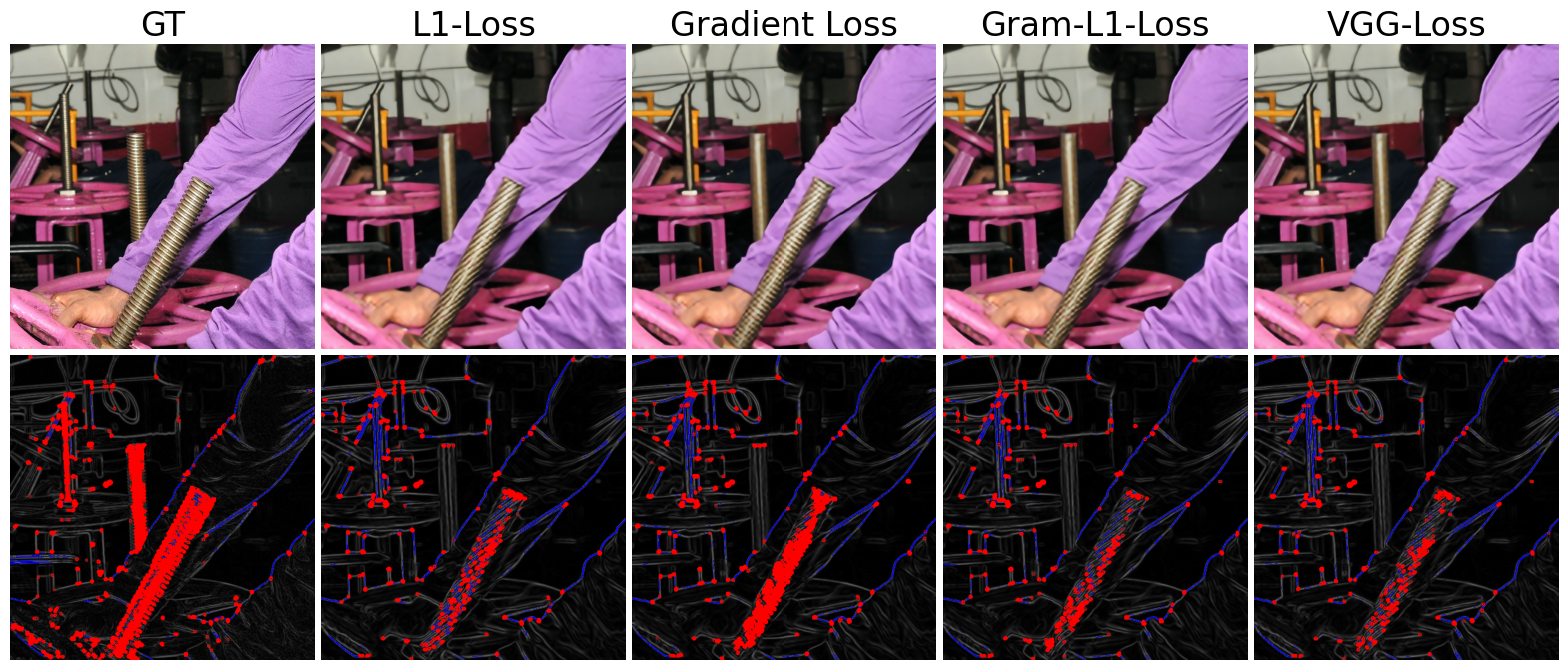}

   \caption{Edge/corner maps for the SRNO model trained with different loss functions on a DIV2K image ($5 \times$ scaling). The Gradient Loss produces sharper edges (blue curves) and corners (red dots), indicating improved texture preservation.}
   \label{fig:divk2kloss}
\end{figure}

\textbf{Impact of Training Objective on Targeted Perceptual Improvements:} We analyze how modifying the loss function affects a model's ability to preserve texture fidelity. Figure~\ref{fig:losschange} reports the change in four texture-sensitive IQA metrics when the standard L1 loss is replaced with more texture-preserving objectives: L1-Gram loss, which augments L1 with a Gram-matrix consistency term to capture texture and style; VGG loss, a feature-space distance computed with a pretrained VGG network; and a hybrid pixel-gradient loss that combines L1 error with first-order gradient consistency. Overall, the metrics respond positively to the modified objectives, with the hybrid pixel-gradient loss showing the most consistent improvements, corroborated qualitatively by sharper edges and more distinct corner structures in the super-resolved images presented in Figure~\ref{fig:divk2kloss}.

\textbf{Hybrid Pixel-Gradient Loss:} The hybrid loss augments the pixel-wise L1 error with a first-order gradient consistency term, encouraging preservation of local edge structure and texture sharpness. The formulation follows prior work leveraging gradient-based constraints for texture guidance. We employ it not as a methodological novelty but as a controlled instrument for isolating the effect of objective design within our benchmark. For ground-truth and predicted images $I$ and $\hat{I}$:

{\small
\begin{equation}
\begin{gathered}
    \mathcal{L} = \lambda_{\mathrm{L1}} \mathcal{L}_{\mathrm{L1}} + \lambda_{\mathrm{grad}} \mathcal{L}_{\mathrm{grad}}, \\
    \mathcal{L}_{L1} = \| \hat{I} - I \|_{1}, \\
    \mathcal{L}_{grad} = \| \nabla_{x} \hat{I} - \nabla_{x} I \|_{1} + \| \nabla_{y} \hat{I} - \nabla_{y} I \|_{1}.
\end{gathered}
\end{equation}
}
$\lambda_{l1}$ and $\lambda_{grad}$ represent the contribution of each term in the final loss, and are set to 1 and 0.05, respectively, in our experiments. Despite its simplicity, the hybrid loss yields consistent gains in texture-sensitive metrics (FSIM, VIF, SR-SIM, GMSD) over standard L1, with corresponding improvements in edge and corner fidelity, as presented in Figure~\ref{fig:divk2kloss}, demonstrating that targeted objective design can deliver measurable perceptual gains even where architectural changes have saturated.

\subsection{Scaling Behavior Analysis}
\label{subsec:scaling}
We investigate three axes of scaling behavior in INR-based ASSR: model capacity, training compute, and data diversity.

\textbf{Model Capacity:} We replace the EDSR~\cite{lim2017enhanced} encoder (1.5Million parameters) with the higher-capacity RDN~\cite{zhang2018residual} encoder (21.9M parameters), keeping all other settings identical. Figure\hyperref[fig:iqa_results]{~\ref*{fig:iqa_results}e)} shows a general improvement across IQA metrics with the larger encoder.

\textbf{Training Compute:} We compare models trained for 100 versus 150 epochs under otherwise identical settings. Figure\hyperref[fig:iqa_results]{~\ref*{fig:iqa_results}f)} shows a general improvement with longer training. Additionally, varying the batch size (8, 16, 32) across all techniques yields a similarly consistent improvement with larger batches (details in Appendix).

\textbf{Data Diversity:} In INR-based ASSR, the effective dataset size is determined not by the number of images but by the number of sampled pixel-coordinate pairs. We therefore study two complementary axes of data scaling, i.e., increasing the patch size, which expands the spatial coordinates available per image and thus the volume of samples observed per epoch, and widening the training scale range from 1--4 to 1--6, which broadens the sampling distribution over scaling factors. Figure\hyperref[fig:iqa_results]{~\ref*{fig:iqa_results}c, d, g)} shows a general trend of modest improvement along both axes.

Across all three axes, performance improves consistently and monotonically, indicating that the scaling trends reported for other deep learning problems~\cite{rang2024empirical, cherti2023reproducible} also manifest in INR-based ASSR. However, the magnitude of improvement diminishes as capacity and compute grow, consistent with the aforementioned architectural saturation observation. We emphasize that these are empirical trends observed over a discrete set of configurations, rather than fitted scaling laws.

%% file: sec/5_conclusion.tex
\section{Conclusion and discussion}
\label{sec:conclusion}
In this study, we presented a unifying and comprehensive framework for an empirical analysis of existing INR-based ASSR methods. Specifically, we conducted a diverse set of experiments to investigate the impact of training configurations, including variations in objective design and optimization strategies, as well as scaling behavior across model capacity, compute, and data diversity. We presented a detailed evaluation of existing methods under diverse training settings, utilizing multiple image quality assessment (IQA) metrics, alongside a qualitative assessment of perceptual improvements achieved through targeted training strategies. An aggregated ranking framework was introduced that provides a robust view of each architecture's relative performance across diverse training settings. Finally, we provide a centralized and reproducible code repository to support future research extensions and promote transparent, fair comparisons among existing and forthcoming techniques.

The key insights resulting from this empirical study include: 1) Existing INR-based methods have largely stabilized, approaching a saturation point, as reflected by the marginal performance gains observed. 2) Training configurations have a significant impact on model performance and cannot be neglected, as careful tuning can lead to meaningful improvements across architectures. 3) Targeted perceptual improvements can be achieved through focused training, as demonstrated by auxiliary loss functions yielding improved texture fidelity. Moreover, architectures differ in their sensitivity to individual IQA metrics, underscoring the limits of PSNR-only evaluation. 4) INR-based ASSR exhibits consistent, monotonic scaling behavior, previously unexamined in this setting, with performance improving with increased model capacity, training compute, and training data diversity, though with diminishing returns as complexity grows.

Finally, based on the conducted experiments and analysis, we identify the following areas for future research:

\begin{itemize}
    \item \textbf{Unified and diverse evaluation benchmarking framework:} Future works can benefit from adopting a unified and diverse benchmarking framework, like the one proposed, that integrates multiple IQA metrics and varied training configurations. This can ensure a more consistent and transparent evaluation for future methods, enabling identification of both general performance trends and model-specific strengths.
    \item \textbf{Need for more complex and diverse benchmark datasets:} The saturation of existing architectures on current benchmarks highlights the need for more comprehensive and diverse datasets. These should include both general and domain-specific collections, annotated for perceptual attributes such as high texture fidelity, consistent colors, strong edge structure, semantic coherence, occlusion handling, and depth complexity. Such datasets would complement existing evaluation metrics and provide a stronger basis for evaluating perceptual gains. Furthermore, extending domain-specific datasets beyond faces and real-scene text would enable future work to evaluate indoor versus outdoor environments, photorealistic versus non-photorealistic images, and other specialized domains, providing deeper insights into model generalization and robustness across diverse visual contexts.
    \item \textbf{Targeted perceptual optimization in ASSR:} The exploration of training strategies and architectural designs targeted at achieving specific perceptual gains remains largely underexplored. The empirical findings from our study, particularly the controlled experiments with the hybrid pixel-gradient loss, demonstrate the potential of such targeted optimization in enhancing perceptual quality. Future studies could focus on developing ASSR methods that explicitly target perceptual attributes such as texture fidelity, edge sharpness, and semantic coherence, through refined architectural components and tailored training configurations. 
\end{itemize}

%% file: sec/X_suppl.tex
\clearpage
\setcounter{page}{1}
\section*{Appendix}
\label{sec:appendix}
This appendix provides additional details and analyses supporting the main paper:
\begin{itemize}
    \item Appendix~\ref{sec:experimental}: Experimental setup, including data preparation, model specifications, optimization settings, and training convergence.
    \item Appendix~\ref{sec:gradloss}: Extended analysis of the hybrid pixel-gradient loss, including qualitative comparisons and the effect of the gradient weighting term $\lambda_{\mathrm{grad}}$.
    \item Appendix~\ref{sec:scalingsup}: Supplementary scaling-behavior results, including the batch-size analysis.
    \item Appendix~\ref{sec:confint}: Confidence-interval analysis across seeds and training recipes.
    \item Appendix~\ref{sec:detailedres}: Detailed per-dataset results, domain-specific evaluations, and an analysis of Y-channel versus RGB evaluation bias.
\end{itemize}

Throughout the appendix, higher values indicate better performance for PSNR, SSIM, FSIM, VIF, and SR-SIM, whereas lower values are preferred for GMSD and LPIPS.

\input{appendix/1_experimental}

\input{appendix/2_gradloss}

\input{appendix/3_scalingsup}

\input{appendix/4_confint}

\input{appendix/5_detailedres}

%% file: appendix/1_experimental.tex
\section{Experimental setup}
\label{sec:experimental}

\subsection{Data preparation} 
\label{sec:dataprep}
For training, 800 images from DIV2K were processed through a data generator that extracted patches of size $48^{2}$, $64^{2}$, or $128^{2}$, depending on the selected configuration. Bicubic downsampling was applied to generate the corresponding low-resolution patches from the high-resolution images, with the target non-integer scaling factor sampled uniformly from a distribution $\mathcal{U}(x,y)$, where $x=1$ and $y$ was set to 4 or 6 based on the selected training configuration. To further increase data diversity, each patch was randomly augmented through horizontal, vertical, or diagonal flips. Each image was repeated 40 times, with a new combination of random augmentation, patch selection, and scaling factor applied at each repetition, creating a highly varied training set. 

For testing, bicubic downsampling was applied to generate low-resolution inputs from the ground-truth high-resolution images of DIV2K, Set5, Set14, BSD100, Urban100, SVT, and CelebA-HQ datasets. No augmentations, random scaling, or patch extraction were applied. The full images were passed through each network as a set of coordinates to obtain the upscaled RGB pixel values.

\subsection{Model specifications} 
\label{sec:modelspec}
Table~\ref{tab:modelspecs} presents details about the parameter counts and memory requirements for each model and encoder used in the empirical study, consistent with previously reported specifications~\cite{jiang2025hiif}. Both EDSR and RDN encoders produce latent feature maps of size $(B, 64, H, W)$, where $64$ is the channel dimension, $H \times W$ corresponds to the input patch size, and $B$ is the batch size. The upsampling modules in both encoders were not used. EDSR employs 16 residual blocks, while RDN uses 16 RDB blocks, following conventions from prior INR-based experiments. The decoder architectures were left unchanged from the original works, ensuring a fair comparison across techniques.

\begin{table}
\centering
\begin{tabular}{lcc}
\toprule
Model & Parameters (M) & Memory (GB) \\
\midrule
MetaSR & 0.45 & 1.2 \\
LIIF   & 0.35 & 1.3 \\
LTE    & 0.49 & 1.4\\
SRNO   & 0.81 & 7.1 \\
CiaoSR & 1.43 & 12.6 \\
HIIF   & 1.33 & 1.5 \\
\midrule
EDSR   & 1.5 & 2.2 \\
RDN   & 21.9 & 3.0 \\
\bottomrule
\end{tabular}
\caption{Parameter counts (in million M) and memory requirements for INR-based ASSR models and encoders.}
\label{tab:modelspecs}
\end{table}

\subsection{Optimization setup} 
\label{sec:optsetup}
All models were trained with a starting learning rate of $4 \times 10^{-4}$ under both the SGDR and multi-step schedules. Adam and AdamW share $\beta_{1}=0.9$, $\beta_{2}=0.999$, and $\varepsilon=10^{-8}$; AdamW additionally uses a weight decay of 0.01. For the multi-step scheduler, the learning rate was decayed every 25 epochs. SGDR used an initial cycle length of 25 epochs and a minimum learning rate of $2 \times 10^{-6}$. Unless otherwise specified, EDSR~\cite{lim2017enhanced} encoder-based experiments used a batch size of 32 and RDN~\cite{zhang2018residual} encoder-based experiments a batch size of 16. The batch-size scaling analysis (Appendix~\ref{sec:scalingsup}) was run as a dedicated set of configurations with batch sizes of 8, 16, and 32.

\subsection{Training loss and convergence} 
\label{sec:lossandconv}
\begin{figure*}
  \centering
   \includegraphics[width=\linewidth]{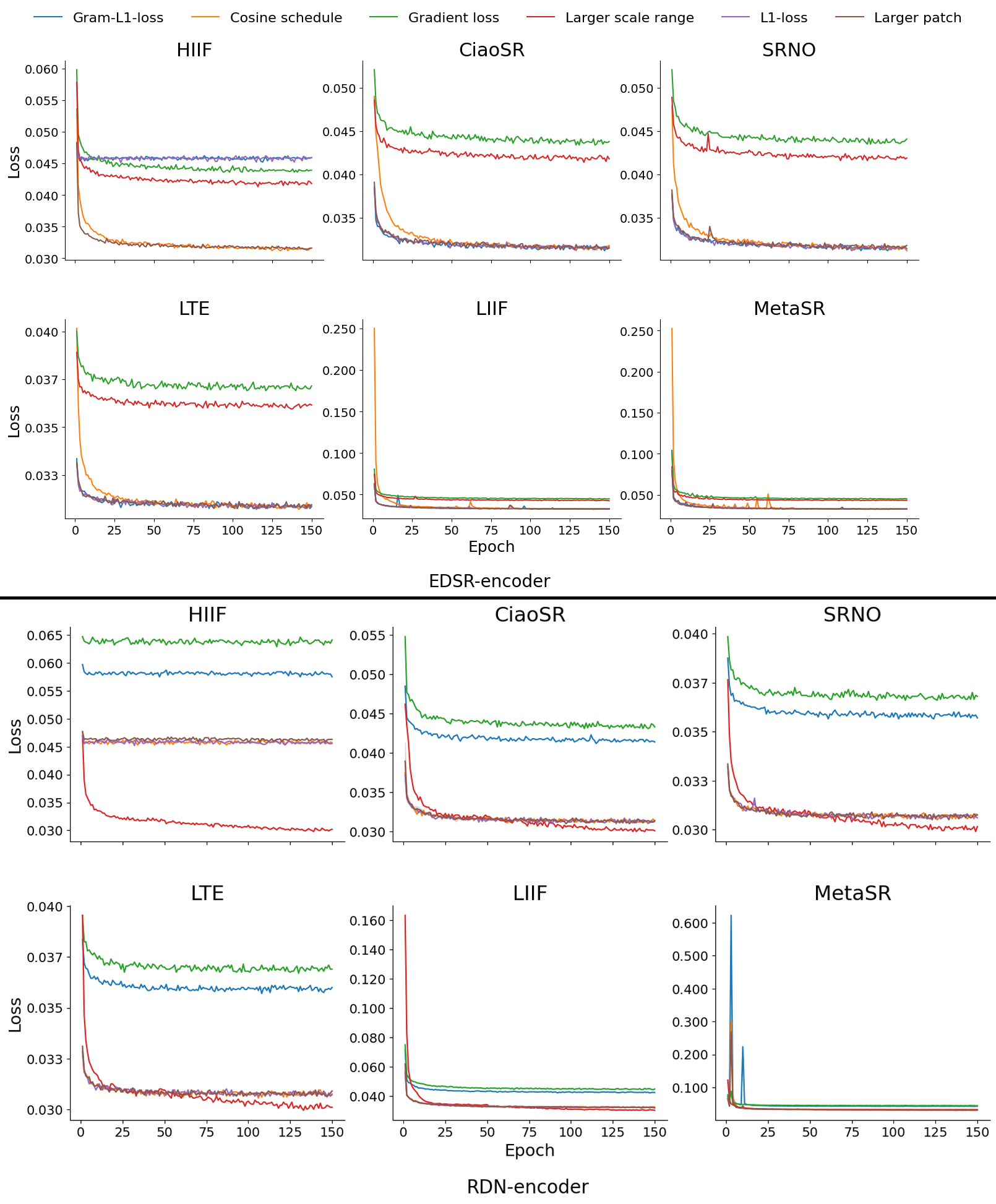}

   \caption{Training losses plotted against the epochs for different techniques trained under different settings. All achieved convergence well before 150 epochs.}
   \label{fig:trainingloss}
\end{figure*}

All models were trained for 150 epochs, with an additional set of experiments using 100 epochs to examine training cost and scaling behavior. Prior works reported results for 1000 epochs, but we limited the training to 150 epochs to balance computational feasibility and convergence. Figure~\ref{fig:trainingloss} presents the training loss curves for various models under different configurations, showing that convergence is generally achieved well before 150 epochs, ensuring fair comparisons across models.

%% file: appendix/2_gradloss.tex
\section{Hybrid pixel-gradient loss analysis}
\label{sec:gradloss}
The main paper (Section~4.3) uses a hybrid L1-gradient objective as a controlled instrument for isolating the effect of objective design, showing consistent gains in texture-sensitive metrics over standard L1. This appendix extends that analysis along three lines: a
quantitative sweep of the gradient-weighting term, qualitative comparisons against baseline objectives, and an analysis of texture retention in the encoder's latent space.

\subsection{Effect of the gradient-weighting term}
\label{sec:lambdasweep}
\begin{figure}[!htb]
  \centering
   \includegraphics[width=\linewidth]{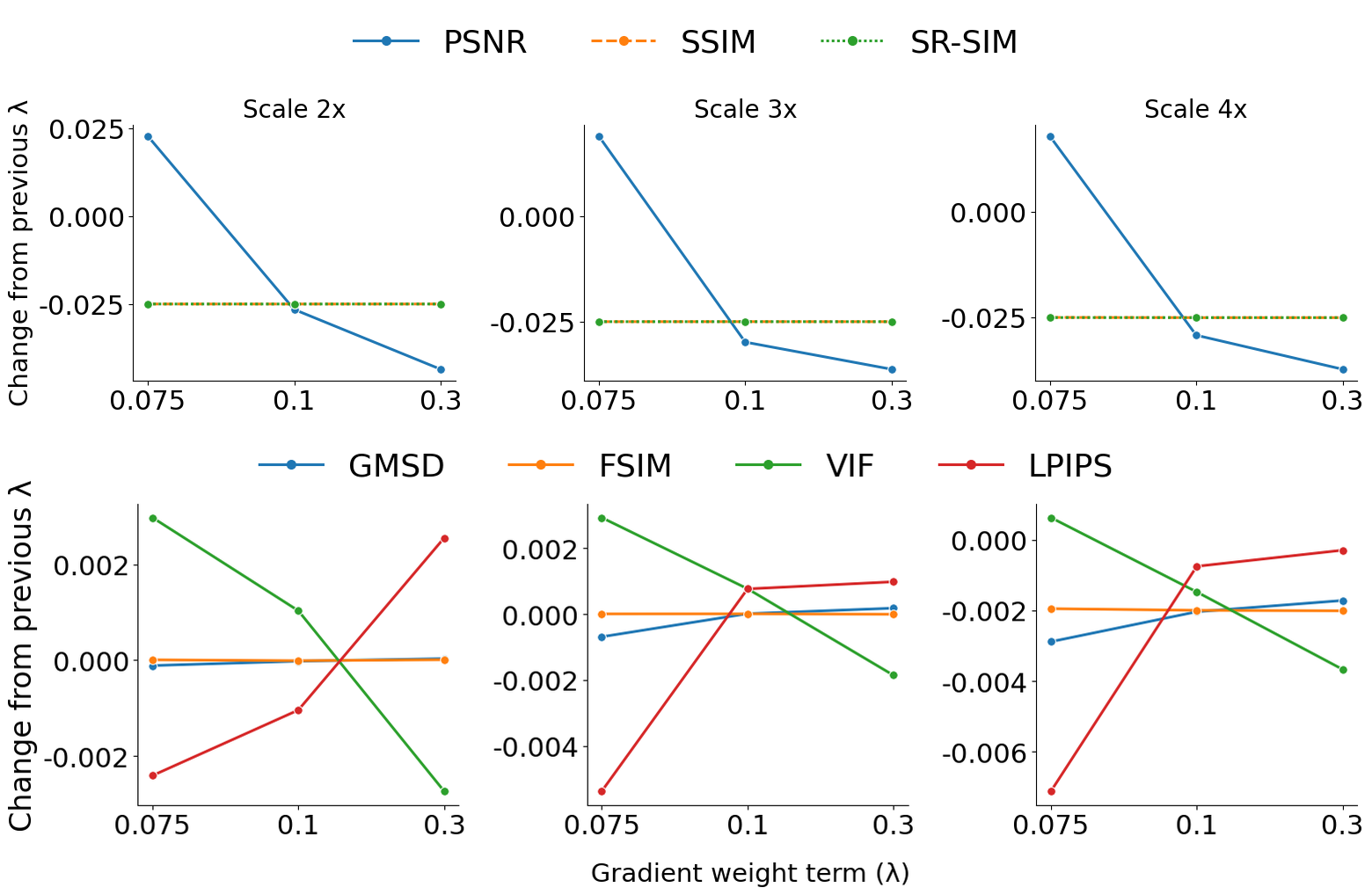}

   \caption{Relative change in various IQA metrics as the weight of the texture fidelity term in the hybrid loss is increased. Results are reported on the DIV2K validation set. The x-axis denotes the updated weight values, while the y-axis shows the change relative to the previous, lower weight.}
   \label{fig:lossweightchange}
\end{figure}

The hybrid L1-gradient objective consistently improves texture-sensitive metrics, enhancing texture preservation in super-resolved images, particularly improving edge and corner fidelity. We further investigate these effects by increasing the weighting of the loss term responsible for texture fidelity. For this analysis, we use the SRNO~\cite{wei2023super} model trained under identical settings, with the gradient consistency term’s weight incrementally raised to 0.075, 0.1, and 0.3. The first part of Figure~\ref{fig:lossweightchange} maps the IQA metrics related to intensity differences, luminance, contrast, and structural similarity, while the second part maps texture-sensitive metrics. Increasing the gradient weight improves texture sharpness up to a certain threshold; beyond this point, the dominant gradient term begins to interfere with the L1 component’s ability to capture pixel-level intensity, resulting in a decline in PSNR and overall reconstruction quality. Table~\ref{tab:bestiqalambda} presents the changes in various IQA metrics for the SRNO model with the EDSR encoder, as the gradient weighting term in the loss function is increased. The results further confirm that moderate gradient weighting consistently improves texture sharpness, whereas overly large weights lead to degradation in pixel-level reconstruction quality.

\begin{table}[!htb]
\centering
\footnotesize
\setlength{\tabcolsep}{3.5pt}
\begin{tabular}{c|ccccccc}
\toprule
$\lambda_{\mathrm{grad}}$ & PSNR & SSIM & GMSD & FSIM & VIF & SR-SIM & LPIPS \\
\midrule
\multicolumn{8}{c}{Scale 2x} \\
\midrule
0.05  & 34.3357 & 0.9991 & 0.0071 & 0.9995 & 0.6204 & 0.9997 & 0.0677 \\
0.075 & \textcolor{red}{34.4314} & 0.9991 & \textcolor{red}{0.0070} & 0.9995 & \textcolor{blue}{0.6234} & 0.9997 & \textcolor{blue}{0.0653} \\
0.1   & \textcolor{blue}{34.4283} & 0.9991 & \textcolor{red}{0.0070} & 0.9995 & \textcolor{red}{0.6245} & 0.9997 & \textcolor{red}{0.0642} \\
0.3   & 34.3911 & 0.9991 & \textcolor{red}{0.0070} & 0.9995 & 0.6217 & 0.9997 & 0.0668 \\
\midrule
\multicolumn{8}{c}{Scale 3x} \\
\midrule
0.05  & 30.6622 & 0.9976 & 0.0380 & 0.9983 & 0.5055 & 0.9991 & 0.1471 \\
0.075 & \textcolor{red}{30.7503} & 0.9976 & \textcolor{red}{0.0373} & 0.9983 & \textcolor{blue}{0.5084} & 0.9991 & \textcolor{red}{0.1417} \\
0.1   & \textcolor{blue}{30.7406} & \textcolor{red}{0.9977} & \textcolor{red}{0.0373} & 0.9983 & \textcolor{red}{0.5092} & 0.9991 & \textcolor{blue}{0.1425} \\
0.3   & 30.7180 & 0.9976 & \textcolor{blue}{0.0375} & 0.9983 & 0.5073 & 0.9991 & 0.1435 \\
\midrule
\multicolumn{8}{c}{Scale 4x} \\
\midrule
0.05  & 28.7522 & 0.9924 & 0.0703 & 0.9955 & 0.4284 & 0.9978 & 0.2155 \\
0.075 & \textcolor{red}{28.8380} & \textcolor{red}{0.9925} & \textcolor{blue}{0.0692} & 0.9955 & \textcolor{blue}{0.4317} & 0.9978 & \textcolor{red}{0.2091} \\
0.1   & \textcolor{blue}{28.8297} & \textcolor{red}{0.9925} & \textcolor{red}{0.0691} & \textcolor{red}{0.9956} & \textcolor{red}{0.4324} & 0.9978 & \textcolor{blue}{0.2107} \\
0.3   & 28.8053 & \textcolor{red}{0.9925} & 0.0695 & 0.9955 & 0.4303 & 0.9978 & 0.2128 \\
\bottomrule
\end{tabular}
\caption{Impact of varying $\lambda_{\mathrm{grad}}$ on different IQA metrics (SRNO, EDSR encoder, DIV2K validation). The best and second-best results per column are colored red and blue, respectively. When multiple configurations tie for the best value at the reported precision, all are colored red. Columns where all configurations are identical at the reported precision (e.g., FSIM and SR-SIM) are left unhighlighted, as these metrics are saturated at these scales.}
\label{tab:bestiqalambda}
\end{table}

\subsection{Qualitative comparisons}
\label{sec:lossqualitative}
\begin{figure}[!htb]
  \centering
   \includegraphics[width=\linewidth]{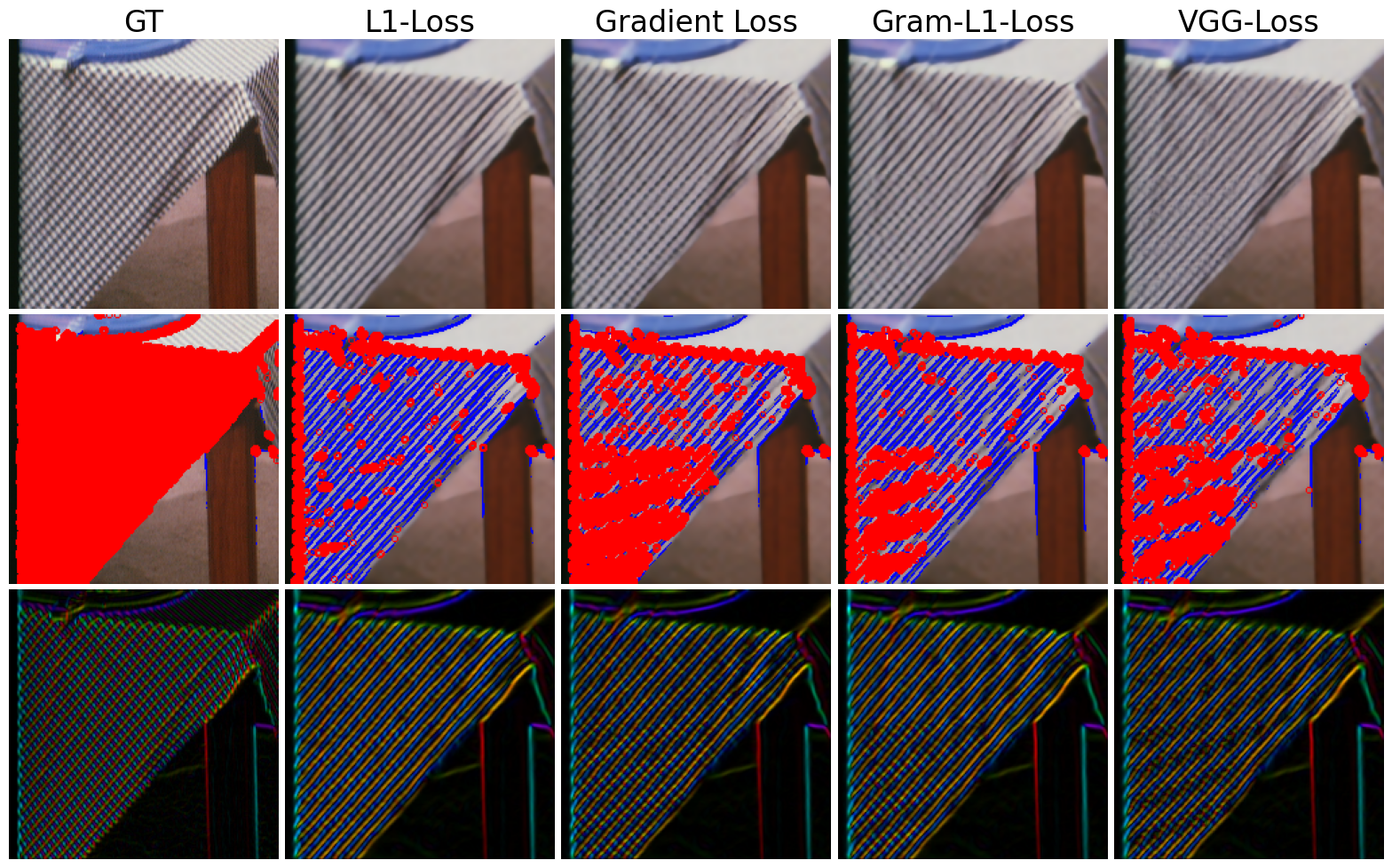}

   \caption{First-order derivative and edge-corner maps for the SRNO model trained with different loss functions on a Set14 image (scaled 4.5x). The Gradient Loss produces sharper edges and corners.}
   \label{fig:set14loss}
\end{figure}

Next, we provide additional qualitative evaluations comparing our loss against baseline objectives and further analyzing the impact of the gradient-weighting factor. Figure~\ref{fig:set14loss} provides a qualitative comparison between the hybrid loss and two baseline objectives. For each loss, we additionally visualize the corresponding edge and corner maps, along with the first-order derivative for the same image patches, to highlight differences in texture fidelity and structural detail.

\begin{figure}[!htb]
  \centering
   \includegraphics[width=\linewidth]{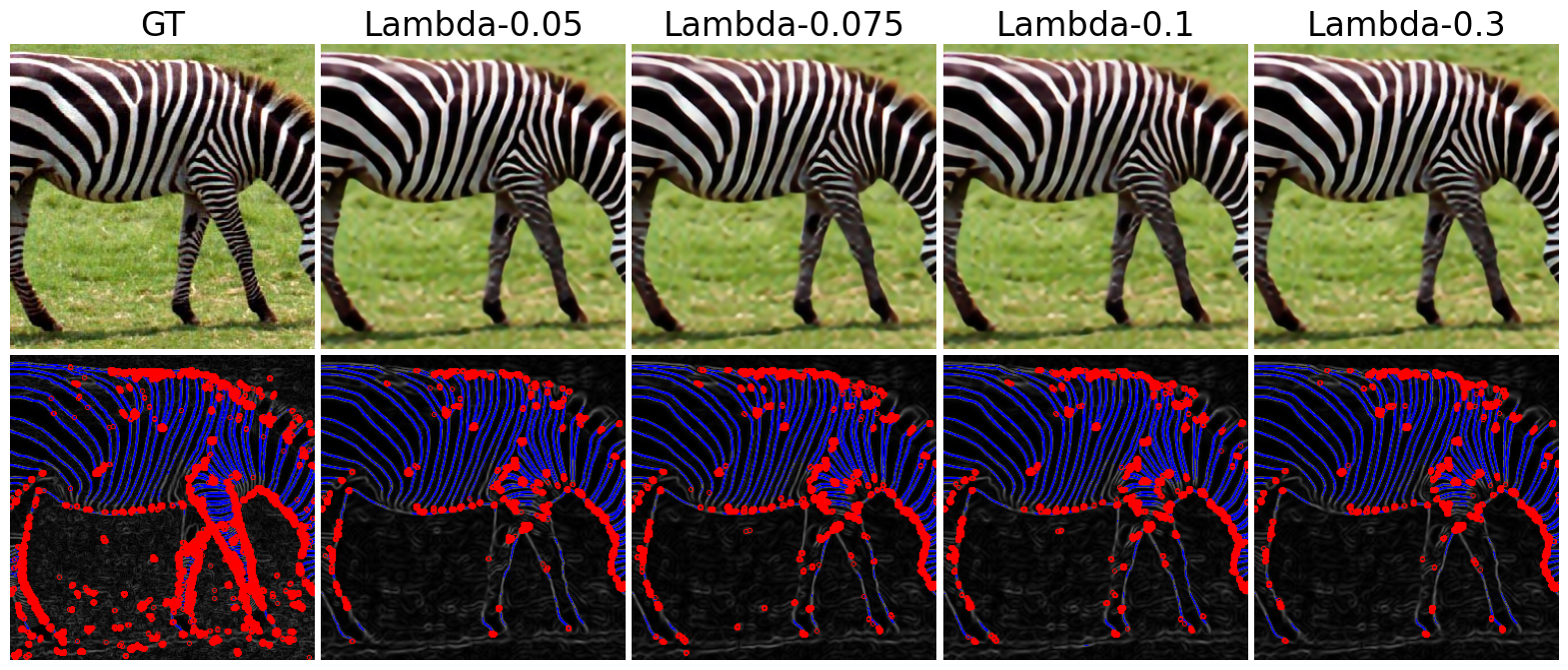}

   \caption{Edge/corner map for image selected from Set14 scaled to 4.5x. A general increase in corners and edges is observed as the lambda is increased.}
   \label{fig:textcomparesupp1}
\end{figure}

Next, we present Figure~\ref{fig:textcomparesupp1}, which shows the edge and corner maps for the same generated image under different values of $\lambda$, illustrating the influence of the gradient weighting term. As observed, increasing $\lambda$ enhances the sharpness and improves edge and corner retention, indicating stronger texture emphasis, consistent with our earlier quantitative observations.

\subsection{Texture retention in the latent space}
\label{sec:featurespace}
\begin{figure*}[!htb]
  \centering
   \includegraphics[width=1\linewidth]{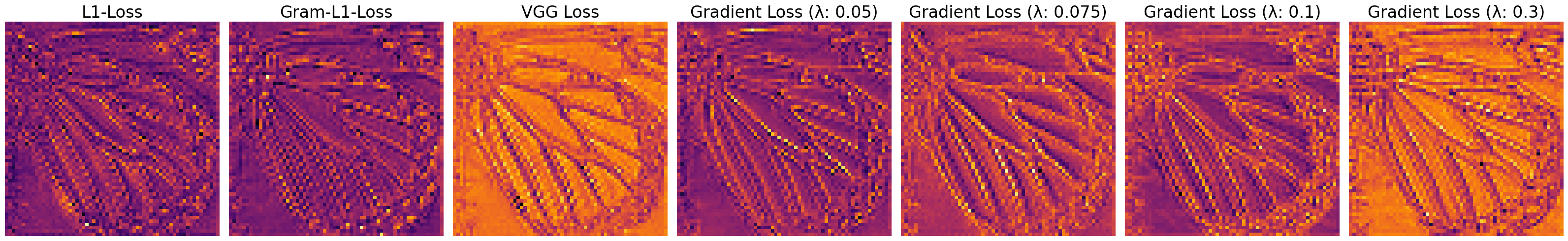}

   \caption{Feature maps extracted from the EDSR encoder of the SRNO model, trained using different loss functions.}
   \label{fig:featuremaps}
\end{figure*}

To further examine texture retention at the feature level, we extract intermediate latent representations from the EDSR encoder for SRNO and compute channel-wise mean feature maps. Figure~\ref{fig:featuremaps} presents a comparison of these feature maps. It can be observed that different objective functions lead to distinct levels of feature smoothness, edge retention, and contrast in the learned representations. Moving left to right, we observe improved structural continuity and sharper texture patterns as $\lambda$ increases. 

We further investigate this by computing the Gray-Level Co-Occurrence Matrix (GLCM)~\cite{haralick2007textural} for the intermediate encoded latent features and report the extracted statistics in Table~\ref{tab:glcmfeature}. The results show that the hybrid loss with moderate values of $\lambda$ maintains high contrast and GLCM dissimilarity, while preserving moderate energy and Angular Second Moment (ASM). This behavior indicates that the model retains more pronounced texture cues in the latent space, while still maintaining meaningful structure without overfitting to pixel-level noise.

\begin{table}[!htb]
\centering
\footnotesize
\setlength{\tabcolsep}{3.5pt}
\begin{tabular}{c|ccccc}
\toprule
Loss & Contrast & Dissimilarity & Energy & Correlation & ASM \\
\midrule
L1-Loss & 1700 & 31 & 0.0155 & 0.0773 & 0.0002 \\
\makecell{Gradient Loss \\($\lambda$: 0.05)} & 1840 & 30 & 0.0167 & 0.0189 & 0.0003 \\
\makecell{Gradient Loss \\($\lambda$: 0.075)} & 1947 & 32 & 0.0153 & 0.0591 & 0.0002 \\
\makecell{Gradient Loss \\($\lambda$: 0.1)} & 2024 & 32 & 0.0153 & 0.0685 & 0.0002 \\
\makecell{Gradient Loss \\($\lambda$: 0.3)} & 1565 & 29 & 0.0153 & 0.3020 & 0.0002 \\
\bottomrule
\end{tabular}

\caption{Texture descriptor statistics from EDSR-SRNO encoder feature maps trained with various lambda values of the hybrid-gradient loss functions.}
\label{tab:glcmfeature}
\end{table}

%% file: appendix/3_scalingsup.tex
\section{Supplementary scaling-behavior analysis}
\label{sec:scalingsup}

\begin{figure}[!htb]
    \centering
    \includegraphics[width=0.95\linewidth]{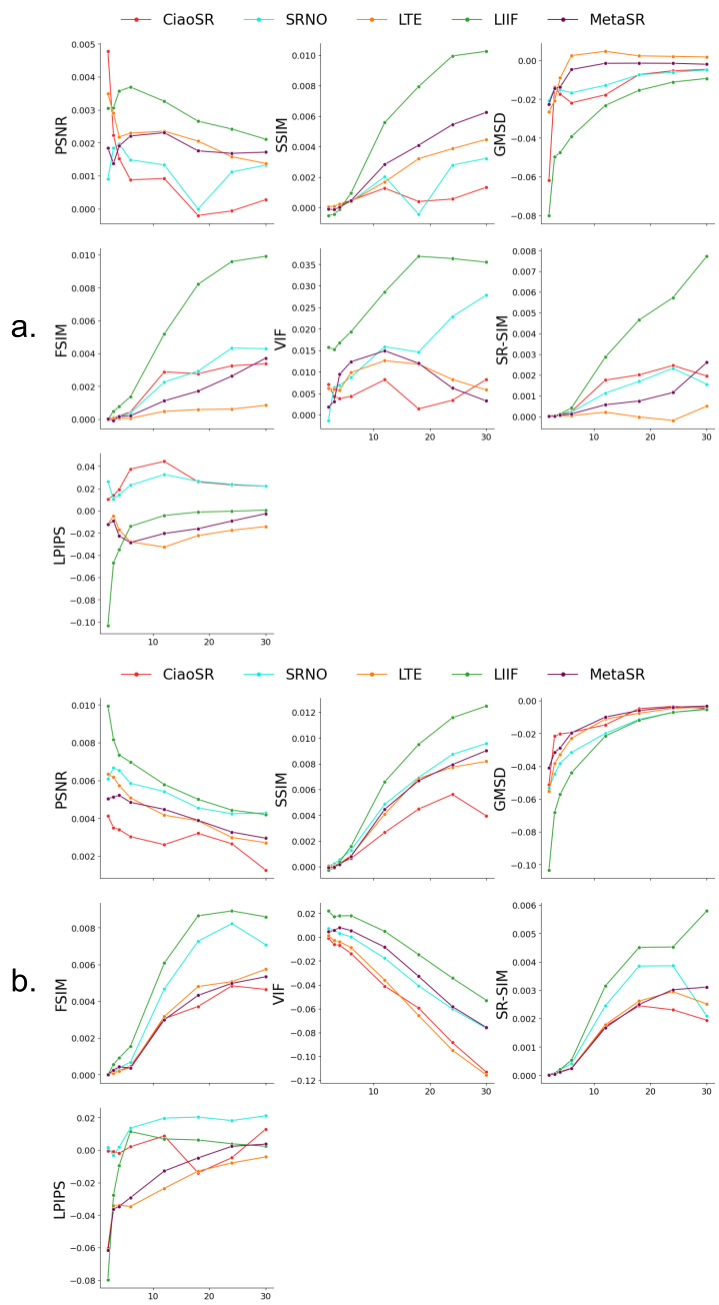}
    \caption{Change in IQA metrics relative to a baseline batch size of 8, across techniques and scales. For PSNR, SSIM, FSIM, VIF, and SR-SIM, higher is better; for GMSD and LPIPS, lower is better. HIIF is excluded due to training instability under the multi-step schedule used in these experiments, which would otherwise obscure general trends. \textbf{(a, b)} Increasing batch size from 8 to 16 (a) and from 8 to 32 (b) yields a consistent improvement trend, with the larger increase reinforcing the positive effect.}
    \label{fig:batch}
\end{figure}

The main paper (Section 4.4) examines scaling behavior along three axes: model capacity, training compute, and data diversity. This appendix provides the supplementary batch-size analysis referenced there, which complements the training-duration comparison as a second axis of the training-compute study. We trained all techniques with batch sizes of 8, 16, and 32 under the L1-Loss recipe with the multi-step scheduler, keeping all other settings identical, including the number of training epochs and the initial learning rate. Figure~\ref{fig:batch} reports the change in each IQA metric relative to the batch-size-8 baseline, plotted against scale. Moving to a higher batch size yields a general improvement across metrics and architectures. HIIF is excluded from this analysis, as it exhibits training instability under the multi-step schedule used in these experiments, which would otherwise obscure the general trends.

%% file: appendix/4_confint.tex
\section{Confidence Intervals}
\label{sec:confint}
We quantify the statistical robustness of the comparisons in the main paper along two axes: variability across training seeds and variability across training recipes. Table~\ref{tab:confidence} reports both analyses at the $12 \times$ scaling factor on DIV2K. Seed-level statistics are computed over 3 independent training runs of the L1-Loss recipe with the multi-step scheduler and the EDSR encoder. Recipe-level statistics are computed over
the 9 training recipes. All intervals are 95\% confidence intervals and values below $5 \times 10^{-4}$ are displayed as 0.000.

\begin{table*}[!htb]
\centering
\begin{tabular}{l|ccccccc}
\toprule
Model & PSNR & SSIM & GMSD & FSIM & VIF & SR-SIM & LPIPS \\
\midrule
\midrule
\multicolumn{8}{c}{Running for 3 different training seeds} \\
\midrule
HIIF & 21.91 ± 0.001 & 0.78 ± 0.001 & 0.24 ± 0.000 & 0.85 ± 0.000 & 0.11 ± 0.000 & 0.91 ± 0.000 & 0.68 ± 0.003 \\
CiaoSR & 23.63 ± 0.003 & 0.88 ± 0.004 & 0.20 ± 0.007 & 0.93 ± 0.006 & 0.17 ± 0.003 & 0.96 ± 0.003 & 0.54 ± 0.039 \\
SRNO & 23.62 ± 0.005 & 0.88 ± 0.002 & 0.20 ± 0.006 & 0.93 ± 0.004 & 0.17 ± 0.001 & 0.96 ± 0.002 & 0.54 ± 0.060 \\
LTE & 23.60 ± 0.000 & 0.88 ± 0.000 & 0.20 ± 0.000 & 0.93 ± 0.000 & 0.17 ± 0.000 & 0.96 ± 0.000 & 0.55 ± 0.000 \\
LIIF & 23.63 ± 0.000 & 0.88 ± 0.000 & 0.20 ± 0.000 & 0.93 ± 0.000 & 0.17 ± 0.000 & 0.96 ± 0.000 & 0.53 ± 0.000 \\
MetaSR & 23.49 ± 0.000 & 0.87 ± 0.000 & 0.20 ± 0.000 & 0.93 ± 0.000 & 0.16 ± 0.000 & 0.96 ± 0.000 & 0.56 ± 0.000 \\

\midrule
\multicolumn{8}{c}{Across different training configurations} \\
\midrule

HIIF & 27.89 ± 0.856 & 0.99 ± 0.006 & 0.09 ± 0.017 & 0.99 ± 0.004 & 0.39 ± 0.039 & 1.00 ± 0.002 & 0.27 ± 0.051 \\
CiaoSR & 28.83 ± 0.025 & 0.99 ± 0.000 & 0.07 ± 0.000 & 1.00 ± 0.000 & 0.43 ± 0.001 & 1.00 ± 0.000 & 0.21 ± 0.002 \\
SRNO & 28.83 ± 0.046 & 0.99 ± 0.000 & 0.07 ± 0.001 & 1.00 ± 0.000 & 0.43 ± 0.002 & 1.00 ± 0.000 & 0.21 ± 0.002 \\
LTE & 28.78 ± 0.064 & 0.99 ± 0.000 & 0.07 ± 0.001 & 1.00 ± 0.000 & 0.43 ± 0.002 & 1.00 ± 0.000 & 0.21 ± 0.002 \\
LIIF & 28.67 ± 0.023 & 0.99 ± 0.000 & 0.07 ± 0.000 & 1.00 ± 0.000 & 0.43 ± 0.001 & 1.00 ± 0.000 & 0.21 ± 0.002 \\
MetaSR & 28.66 ± 0.027 & 0.99 ± 0.000 & 0.07 ± 0.000 & 0.99 ± 0.000 & 0.43 ± 0.001 & 1.00 ± 0.000 & 0.21 ± 0.001 \\

\bottomrule
\end{tabular}
\caption{Mean performance and corresponding 95\% confidence intervals (CI) computed across different training seeds and training recipes for each ASSR method at the $12 \times$ scaling factor for DIV2K. Top block: variability across 3 training seeds (L1-Loss recipe, multi-step scheduler, EDSR encoder). Bottom block: variability across the nine training recipes.}
\label{tab:confidence}
\end{table*}

Seed-level variability is negligible for all methods, well below the recipe-level spread, confirming that the ranking differences discussed in the main paper are not artifacts of training stochasticity. Note that HIIF's lower seed-level mean reflects the multi-step scheduler used in this recipe, under which HIIF is known to underperform, and its seed-level variance is nevertheless as small as that of the other methods. Its wide recipe-level intervals likewise stem from this single scheduler-induced failure mode rather than general training instability.

%% file: appendix/5_detailedres.tex
\section{Detailed results}
\label{sec:detailedres}
This section presents an alternative numeric representation of the results. Table~\ref{tab:margimpedsr} reports the PSNR gains of the best-performing models across different scales on the DIV2K validation set. The maximum gain of 0.035dB was achieved by the SRNO model trained with the EDSR-encoder.

\begin{table}[!htb]
\centering
\small
\setlength{\tabcolsep}{4pt}
\begin{tabular}{c|ccc}
\toprule
Scale & Best Model & Second Best Model & Gain in PSNR (dB) \\
\midrule
 & \multicolumn{3}{c}{EDSR-encoder} \\
\midrule
2 & SRNO & HIIF & 0.027 \\
3 & SRNO & LTE & 0.026 \\
4 & SRNO & CiaoSR & \textbf{0.035} \\
6 & SRNO & CiaoSR & 0.007 \\
12 & SRNO & CiaoSR & 0.007 \\
18 & SRNO & CiaoSR & 0.015 \\
24 & SRNO & LTE & 0.014 \\
30 & SRNO & LTE & 0.009 \\

\midrule
 & \multicolumn{3}{c}{RDN-encoder} \\
\midrule
2 & SRNO & LTE & 0.020 \\
3 & SRNO & LTE & 0.014 \\
4 & SRNO & CiaoSR & \textbf{0.022} \\
6 & CiaoSR & SRNO & 0.004 \\
12 & SRNO & CiaoSR & 0.001 \\
18 & CiaoSR & SRNO & 0.002 \\
24 & SRNO & LTE & 0.012 \\
30 & SRNO & LTE & 0.016 \\

\bottomrule
\end{tabular}
\caption{PSNR gains of the best-performing models at various scales for EDSR and RDN-based encoders. The largest gain for each encoder is highlighted in bold.}
\label{tab:margimpedsr}
\end{table}

\subsection{Per-metric top performers}
\label{sec:topperf}
Next, we present a breakdown of the previous rankings of each model across different training configurations. We averaged the performance of each technique across different settings to provide mean scores per IQA metric. The top-performing models were then identified for each metric, scale, dataset, and encoder, and are presented in Tables~\ref {tab:topresults1} and ~\ref{tab:topresults2}. Table~\ref{tab:topresults1} summarizes the best-performing models for DIV2K, Set5, and Set14, while Table~\ref{tab:topresults2} reports the corresponding results for BSD100 and Urban100.

\begin{table*}[!htb]
\centering
\footnotesize
\setlength{\tabcolsep}{4pt}
\begin{tabular}{cccccccc}
\toprule
Scale & PSNR & SSIM & GMSD & FSIM & VIF & SR-SIM & LPIPS \\
\midrule
\multicolumn{8}{c}{DIV2K} \\
\midrule
\multicolumn{8}{c}{EDSR-encoder} \\
\midrule
2 & SRNO: 34.4143 & LTE: 0.9991 & LTE: 0.0069 & LTE: 0.9995 & SRNO: 0.6229 & LTE: 0.9997 & CiaoSR: 0.0643 \\
3 & SRNO: 30.7396 & LTE: 0.9976 & SRNO: 0.0374 & LTE: 0.9983 & LTE: 0.5084 & MetaSR: 0.9991 & CiaoSR: 0.1407 \\
4 & CiaoSR: 28.8297 & SRNO: 0.9924 & SRNO: 0.0694 & SRNO: 0.9955 & CiaoSR: 0.4318 & SRNO: 0.9978 & CiaoSR: 0.2085 \\
6 & CiaoSR: 26.6362 & SRNO: 0.9758 & SRNO: 0.1219 & LTE: 0.9894 & CiaoSR: 0.3201 & SRNO: 0.9939 & CiaoSR: 0.3263 \\
12 & CiaoSR: 23.6416 & SRNO: 0.8781 & SRNO: 0.1991 & SRNO: 0.929 & CiaoSR: 0.1673 & SRNO: 0.9596 & CiaoSR: 0.5384 \\
18 & SRNO: 22.1178 & SRNO: 0.7766 & MetaSR: 0.2326 & SRNO: 0.8613 & SRNO: 0.1124 & SRNO: 0.9196 & CiaoSR: 0.6187 \\
24 & SRNO: 21.1366 & SRNO: 0.6898 & MetaSR: 0.2521 & CiaoSR: 0.8045 & SRNO: 0.0854 & SRNO: 0.8832 & SRNO: 0.6556 \\
30 & SRNO: 20.4501 & SRNO: 0.6253 & CiaoSR: 0.2645 & SRNO: 0.7592 & SRNO: 0.07 & SRNO: 0.8452 & SRNO: 0.6718 \\

\midrule
\multicolumn{8}{c}{RDN-encoder} \\
\midrule
2 & SRNO: 34.5854 & LTE: 0.9992 & LTE: 0.0066 & LTE: 0.9995 & SRNO: 0.6278 & LTE: 0.9997 & SRNO: 0.0629 \\
3 & LTE: 30.8858 & LTE: 0.9977 & LTE: 0.0361 & LTE: 0.9984 & LTE: 0.5141 & LTE: 0.9991 & CiaoSR: 0.1384 \\
4 & SRNO: 28.9513 & LTE: 0.9927 & SRNO: 0.0675 & LTE: 0.9957 & LTE: 0.4368 & LTE: 0.9979 & CiaoSR: 0.2051 \\
6 & CiaoSR: 26.7378 & SRNO: 0.9767 & SRNO: 0.1196 & LTE: 0.9899 & CiaoSR: 0.325 & LTE: 0.9941 & LIIF: 0.321 \\
12 & CiaoSR: 23.7185 & SRNO: 0.8811 & SRNO: 0.1972 & SRNO: 0.9314 & CiaoSR: 0.1707 & SRNO: 0.961 & LIIF: 0.5283 \\
18 & SRNO: 22.1842 & CiaoSR: 0.7808 & CiaoSR: 0.2318 & SRNO: 0.8644 & SRNO: 0.1149 & SRNO: 0.9214 & LIIF: 0.6091 \\
24 & SRNO: 21.1929 & SRNO: 0.6947 & CiaoSR: 0.2513 & CiaoSR: 0.808 & LTE: 0.0872 & CiaoSR: 0.8855 & LIIF: 0.6497 \\
30 & SRNO: 20.4974 & SRNO: 0.6297 & CiaoSR: 0.2636 & SRNO: 0.7626 & LTE: 0.0714 & SRNO: 0.8474 & SRNO: 0.6698 \\
\midrule
\midrule

\multicolumn{8}{c}{Set5} \\
\midrule
\multicolumn{8}{c}{EDSR-encoder} \\
\midrule
2 & SRNO: 37.813 & LTE: 0.9661 & LTE: 0.0021 & SRNO: 0.9789 & SRNO: 0.7261 & SRNO: 0.991 & CiaoSR: 0.0535 \\
3 & SRNO: 34.1995 & LTE: 0.9355 & LTE: 0.0188 & LTE: 0.9518 & LTE: 0.6284 & SRNO: 0.978 & LIIF: 0.1122 \\
4 & CiaoSR: 32.0539 & CiaoSR: 0.9056 & SRNO: 0.0439 & CiaoSR: 0.9273 & CiaoSR: 0.555 & SRNO: 0.9658 & MetaSR: 0.1531 \\
6 & SRNO: 28.7492 & SRNO: 0.8393 & SRNO: 0.0963 & SRNO: 0.8761 & SRNO: 0.4259 & SRNO: 0.9297 & SRNO: 0.2394 \\
8 & LTE: 26.7686 & SRNO: 0.7765 & SRNO: 0.1388 & SRNO: 0.828 & SRNO: 0.3324 & SRNO: 0.8924 & SRNO: 0.3169 \\
12 & SRNO: 24.5048 & SRNO: 0.6882 & SRNO: 0.1941 & SRNO: 0.7631 & SRNO: 0.2263 & MetaSR: 0.8366 & SRNO: 0.4316 \\

\midrule
\multicolumn{8}{c}{RDN-encoder} \\
\midrule
2 & LTE: 37.9457 & SRNO: 0.9665 & LTE: 0.002 & SRNO: 0.9795 & LTE: 0.7292 & SRNO: 0.9913 & SRNO: 0.0527 \\
3 & LTE: 34.3437 & LTE: 0.9369 & SRNO: 0.0182 & LTE: 0.9532 & LTE: 0.6331 & SRNO: 0.9787 & LIIF: 0.1115 \\
4 & CiaoSR: 32.1847 & LTE: 0.9075 & LTE: 0.0425 & CiaoSR: 0.9292 & LTE: 0.5605 & SRNO: 0.9669 & SRNO: 0.1528 \\
6 & SRNO: 28.8677 & LTE: 0.8422 & SRNO: 0.0942 & SRNO: 0.8781 & LTE: 0.4327 & LTE: 0.931 & SRNO: 0.2362 \\
8 & LTE: 26.9388 & LTE: 0.7815 & SRNO: 0.1369 & LTE: 0.8311 & LTE: 0.3413 & SRNO: 0.8942 & SRNO: 0.3116 \\
12 & LTE: 24.5965 & LTE: 0.6915 & SRNO: 0.1925 & LTE: 0.7644 & LTE: 0.2312 & SRNO: 0.8362 & SRNO: 0.4235 \\
\midrule
\midrule

\multicolumn{8}{c}{Set14} \\
\midrule
\multicolumn{8}{c}{EDSR-encoder} \\
\midrule
2 & SRNO: 33.4291 & LTE: 0.9691 & LTE: 0.0079 & LTE: 0.9813 & LTE: 0.6102 & LTE: 0.9917 & CiaoSR: 0.078 \\
3 & SRNO: 30.2015 & LTE: 0.9073 & SRNO: 0.0395 & LTE: 0.9491 & LTE: 0.5047 & LTE: 0.9762 & CiaoSR: 0.1657 \\
4 & SRNO: 28.4768 & CiaoSR: 0.8493 & CiaoSR: 0.0714 & CiaoSR: 0.9089 & LTE: 0.4319 & CiaoSR: 0.9563 & CiaoSR: 0.2375 \\
6 & CiaoSR: 26.314 & CiaoSR: 0.7639 & LTE: 0.1265 & CiaoSR: 0.8433 & CiaoSR: 0.3228 & CiaoSR: 0.9164 & CiaoSR: 0.3659 \\
8 & SRNO: 24.8267 & SRNO: 0.6989 & CiaoSR: 0.1652 & CiaoSR: 0.7896 & CiaoSR: 0.248 & CiaoSR: 0.8813 & SRNO: 0.4676 \\
12 & SRNO: 23.0719 & SRNO: 0.6104 & CiaoSR: 0.209 & SRNO: 0.7133 & LTE: 0.1639 & SRNO: 0.8168 & SRNO: 0.5713 \\
\midrule
\multicolumn{8}{c}{RDN-encoder} \\
\midrule
2 & LTE: 33.5897 & LTE: 0.9696 & LTE: 0.0076 & LTE: 0.9819 & LTE: 0.6152 & LTE: 0.9919 & CiaoSR: 0.0768 \\
3 & LTE: 30.3271 & LTE: 0.9095 & LTE: 0.0384 & LTE: 0.9504 & LTE: 0.5107 & LTE: 0.9769 & CiaoSR: 0.163 \\
4 & SRNO: 28.5864 & LTE: 0.8514 & LTE: 0.0696 & CiaoSR: 0.9107 & LTE: 0.4379 & CiaoSR: 0.9573 & CiaoSR: 0.2349 \\
6 & CiaoSR: 26.4213 & LTE: 0.7671 & CiaoSR: 0.1243 & CiaoSR: 0.8463 & LTE: 0.3284 & CiaoSR: 0.9179 & CiaoSR: 0.362 \\
8 & CiaoSR: 24.9101 & LTE: 0.7026 & CiaoSR: 0.163 & CiaoSR: 0.7928 & LTE: 0.2532 & CiaoSR: 0.8824 & SRNO: 0.4613 \\
12 & LTE: 23.1313 & LTE: 0.6144 & CiaoSR: 0.2074 & LTE: 0.7158 & LTE: 0.1679 & SRNO: 0.8186 & SRNO: 0.5663 \\

\midrule
\bottomrule
\end{tabular}
\caption{Top performing model on DIV2K, Set5, and Set14 validation sets, for different scaling factors and different IQA metrics.}
\scriptsize
\label{tab:topresults1}
\end{table*}

\begin{table*}[!htb]
\centering
\footnotesize
\setlength{\tabcolsep}{4pt}

\begin{tabular}{cccccccc}
\toprule
Scale & PSNR & SSIM & GMSD & FSIM & VIF & SR-SIM & LPIPS \\
\midrule
\multicolumn{8}{c}{BSD100} \\
\midrule
\multicolumn{8}{c}{EDSR-encoder} \\
\midrule
2 & LTE: 32.0489 & LTE: 0.8979 & LTE: 0.0087 & LTE: 0.9447 & LTE: 0.5586 & LTE: 0.9716 & CiaoSR: 0.084 \\
3 & SRNO: 28.9941 & LTE: 0.8031 & LTE: 0.0461 & LTE: 0.88 & LTE: 0.4473 & LTE: 0.9368 & CiaoSR: 0.1877 \\
4 & CiaoSR: 27.4932 & CiaoSR: 0.7333 & LTE: 0.0802 & LTE: 0.8261 & LTE: 0.373 & LTE: 0.9091 & CiaoSR: 0.2737 \\
6 & CiaoSR: 25.7802 & CiaoSR: 0.645 & SRNO: 0.1337 & CiaoSR: 0.7464 & CiaoSR: 0.2703 & CiaoSR: 0.8513 & CiaoSR: 0.4157 \\
8 & SRNO: 24.7348 & SRNO: 0.5922 & SRNO: 0.1678 & CiaoSR: 0.6886 & SRNO: 0.2041 & CiaoSR: 0.799 & SRNO: 0.5431 \\
12 & SRNO: 23.4431 & SRNO: 0.5376 & LIIF: 0.2062 & MetaSR: 0.6196 & SRNO: 0.1349 & MetaSR: 0.7351 & SRNO: 0.6639 \\
\midrule
\multicolumn{8}{c}{RDN-encoder} \\
\midrule
2 & LTE: 32.1524 & LTE: 0.8991 & LTE: 0.0085 & SRNO: 0.9458 & LTE: 0.562 & LTE: 0.9722 & CiaoSR: 0.0828 \\
3 & LTE: 29.0817 & LTE: 0.8055 & LTE: 0.0452 & LTE: 0.8817 & LTE: 0.4511 & LTE: 0.9378 & CiaoSR: 0.1855 \\
4 & LTE: 27.5659 & LTE: 0.7359 & LTE: 0.0789 & LTE: 0.8283 & LTE: 0.3771 & LTE: 0.9103 & SRNO: 0.2705 \\
6 & SRNO: 25.8323 & CiaoSR: 0.6473 & SRNO: 0.1322 & CiaoSR: 0.7488 & CiaoSR: 0.2734 & CiaoSR: 0.8529 & SRNO: 0.4111 \\
8 & SRNO: 24.7912 & SRNO: 0.5945 & SRNO: 0.1662 & CiaoSR: 0.6909 & SRNO: 0.2073 & CiaoSR: 0.8006 & SRNO: 0.5361 \\
12 & SRNO: 23.4863 & SRNO: 0.5393 & CiaoSR: 0.2051 & SRNO: 0.6209 & SRNO: 0.1371 & SRNO: 0.7325 & SRNO: 0.6588 \\

\midrule
\midrule

\multicolumn{8}{c}{Urban100} \\
\midrule
\multicolumn{8}{c}{EDSR-encoder} \\
\midrule
2 & SRNO: 31.6894 & SRNO: 0.9961 & SRNO: 0.014 & LTE: 0.9971 & SRNO: 0.6076 & SRNO: 0.9986 & CiaoSR: 0.0563 \\
3 & CiaoSR: 27.8731 & LTE: 0.9909 & CiaoSR: 0.0511 & LTE: 0.9936 & SRNO: 0.4901 & LTE: 0.9949 & CiaoSR: 0.1254 \\
4 & CiaoSR: 25.9269 & CiaoSR: 0.9585 & CiaoSR: 0.0888 & CiaoSR: 0.9812 & CiaoSR: 0.4131 & CiaoSR: 0.9905 & CiaoSR: 0.1918 \\
6 & CiaoSR: 23.659 & CiaoSR: 0.8948 & CiaoSR: 0.1548 & CiaoSR: 0.9424 & CiaoSR: 0.2978 & CiaoSR: 0.9711 & CiaoSR: 0.308 \\
8 & CiaoSR: 22.3679 & CiaoSR: 0.8254 & SRNO: 0.1993 & CiaoSR: 0.8917 & CiaoSR: 0.224 & CiaoSR: 0.9439 & CiaoSR: 0.4077 \\
12 & CiaoSR: 20.8334 & CiaoSR: 0.702 & MetaSR: 0.2502 & CiaoSR: 0.8003 & CiaoSR: 0.1398 & CiaoSR: 0.8947 & CiaoSR: 0.5502 \\

\midrule
\multicolumn{8}{c}{RDN-encoder} \\
\midrule
2 & LTE: 32.0783 & LTE: 0.9964 & LTE: 0.013 & LTE: 0.9975 & LTE: 0.6184 & LTE: 0.9987 & SRNO: 0.0539 \\
3 & LTE: 28.1749 & LTE: 0.9917 & LTE: 0.0481 & LTE: 0.9942 & LTE: 0.5025 & LTE: 0.9953 & CiaoSR: 0.121 \\
4 & CiaoSR: 26.1618 & LTE: 0.9612 & LTE: 0.0848 & LTE: 0.9828 & LTE: 0.4224 & LTE: 0.9912 & CiaoSR: 0.1852 \\
6 & CiaoSR: 23.8322 & LTE: 0.9001 & CiaoSR: 0.1502 & LTE: 0.9462 & CiaoSR: 0.3063 & LTE: 0.973 & CiaoSR: 0.2977 \\
8 & CiaoSR: 22.5116 & CiaoSR: 0.833 & SRNO: 0.195 & SRNO: 0.897 & CiaoSR: 0.2316 & SRNO: 0.9468 & CiaoSR: 0.3955 \\
12 & CiaoSR: 20.9493 & CiaoSR: 0.7117 & CiaoSR: 0.2471 & CiaoSR: 0.8074 & CiaoSR: 0.1456 & CiaoSR: 0.8986 & CiaoSR: 0.5382 \\

\midrule
\bottomrule
\end{tabular}
\caption{Top performing model on BSD100 and Urban100 validation sets, for different scaling factors and different IQA metrics.}
\label{tab:topresults2}
\end{table*}

\subsection{Domain specific results}
\label{sec:domspecres}
The SVT and CelebA-HQ datasets were selected to evaluate the techniques on two targeted domains, i.e., real-scene text and face super-resolution. Similar to the previous tables, aggregated results are reported in Tables~\ref{tab:domainres1} and ~\ref{tab:domainres2}, but here the top three models are highlighted for selected IQA metrics across different scales and encoders.

\begin{table*}[!htb]
\centering
\footnotesize
\setlength{\tabcolsep}{2pt}
\begin{tabular}{c|cccc|cccc}
\toprule
& \multicolumn{4}{c}{EDSR-encoder} & \multicolumn{4}{c}{RDN-encoder} \\
\midrule
Scale & PSNR & SSIM & GMSD & VIF & PSNR & SSIM & GMSD & VIF \\
\midrule
 & CiaoSR: 39.2331 & CiaoSR: 0.9995 & CiaoSR: 0.0038 & CiaoSR: 0.6844 & SRNO: 39.4012 & LTE: 0.9996 & LTE: 0.0036 & SRNO: 0.6895 \\
2 & LIIF: 38.9758 & LIIF: 0.9988 & LIIF: 0.004 & LIIF: 0.681 & LTE: 39.3968 & SRNO: 0.9996 & SRNO: 0.0036 & LTE: 0.6882 \\
 & LTE: 37.7051 & LTE: 0.9984 & LTE: 0.0073 & LTE: 0.6365 & CiaoSR: 39.2945 & CiaoSR: 0.9996 & CiaoSR: 0.0037 & CiaoSR: 0.686 \\

\midrule
 
 & CiaoSR: 35.8452 & CiaoSR: 0.9967 & CiaoSR: 0.0234 & CiaoSR: 0.5496 & SRNO: 35.916 & LTE: 0.9968 & LTE: 0.023 & LTE: 0.552 \\
3 & LIIF: 35.6992 & SRNO: 0.9959 & LIIF: 0.0235 & LIIF: 0.5468 & LTE: 35.9146 & SRNO: 0.9968 & SRNO: 0.023 & SRNO: 0.5518 \\
 & SRNO: 35.0923 & LIIF: 0.9959 & SRNO: 0.0259 & SRNO: 0.523 & CiaoSR: 35.8909 & CiaoSR: 0.9967 & CiaoSR: 0.0231 & CiaoSR: 0.551 \\

\midrule
 
 & CiaoSR: 34.3028 & LTE: 0.9966 & CiaoSR: 0.0439 & CiaoSR: 0.4742 & SRNO: 34.35 & SRNO: 0.9966 & SRNO: 0.0433 & SRNO: 0.4763 \\
4 & LIIF: 34.1901 & SRNO: 0.9965 & LIIF: 0.0445 & LIIF: 0.4714 & LTE: 34.3482 & LTE: 0.9966 & LTE: 0.0435 & LTE: 0.4761 \\
 & LTE: 34.1168 & CiaoSR: 0.9965 & LTE: 0.0446 & LTE: 0.4685 & CiaoSR: 34.3402 & CiaoSR: 0.9966 & CiaoSR: 0.0435 & CiaoSR: 0.4758 \\

\midrule
 
 & CiaoSR: 32.6996 & CiaoSR: 0.9734 & CiaoSR: 0.0801 & CiaoSR: 0.3777 & SRNO: 32.7428 & SRNO: 0.9738 & SRNO: 0.0792 & SRNO: 0.3799 \\
6 & LTE: 32.6244 & LTE: 0.9728 & LTE: 0.0803 & LTE: 0.3756 & CiaoSR: 32.7385 & LTE: 0.9738 & CiaoSR: 0.0793 & LTE: 0.3798 \\
 & SRNO: 32.6226 & SRNO: 0.9727 & SRNO: 0.0803 & LIIF: 0.3752 & LTE: 32.7361 & CiaoSR: 0.9737 & LTE: 0.0797 & CiaoSR: 0.3797 \\

\midrule
 
 & CiaoSR: 31.7457 & CiaoSR: 0.9545 & SRNO: 0.1037 & CiaoSR: 0.3183 & SRNO: 31.7935 & SRNO: 0.9552 & SRNO: 0.1025 & LTE: 0.3207 \\
8 & SRNO: 31.7124 & LTE: 0.9544 & CiaoSR: 0.1037 & LTE: 0.3164 & LTE: 31.7856 & LTE: 0.9551 & CiaoSR: 0.1028 & SRNO: 0.3206 \\
 & LTE: 31.7058 & SRNO: 0.9543 & LTE: 0.104 & SRNO: 0.3161 & CiaoSR: 31.7851 & CiaoSR: 0.9551 & LTE: 0.1032 & CiaoSR: 0.3204 \\

\midrule
 
 & CiaoSR: 30.4292 & CiaoSR: 0.9182 & SRNO: 0.1316 & CiaoSR: 0.2483 & CiaoSR: 30.4741 & CiaoSR: 0.9192 & SRNO: 0.1306 & CiaoSR: 0.2505 \\
12 & SRNO: 30.4262 & SRNO: 0.9179 & CiaoSR: 0.1317 & SRNO: 0.2477 & SRNO: 30.473 & SRNO: 0.9191 & CiaoSR: 0.1307 & SRNO: 0.2503 \\
 & LTE: 30.4086 & LTE: 0.9178 & LTE: 0.1325 & LTE: 0.2474 & LTE: 30.4675 & LTE: 0.919 & LTE: 0.1313 & LTE: 0.2503 \\
\bottomrule
\end{tabular}
\caption{Summary of results achieved on the face domain-specific CelebHQ-A dataset.}
\label{tab:domainres1}
\end{table*}

\begin{table*}[!htb]
\centering
\footnotesize
\setlength{\tabcolsep}{2pt}
\begin{tabular}{c|cccc|cccc}
\toprule
& \multicolumn{4}{c}{EDSR-encoder} & \multicolumn{4}{c}{RDN-encoder} \\
\midrule
Scale & PSNR & SSIM & GMSD & VIF & PSNR & SSIM & GMSD & VIF \\
\midrule

 & LIIF: 45.2465 & CiaoSR: 0.9996 & LIIF: 0.0021 & CiaoSR: 0.8792 & CiaoSR: 44.8995 & SRNO: 0.9996 & LTE: 0.0019 & SRNO: 0.8824 \\
2 & CiaoSR: 44.7078 & LIIF: 0.9995 & CiaoSR: 0.0023 & LIIF: 0.8738 & LTE: 44.7038 & LTE: 0.9996 & CiaoSR: 0.002 & CiaoSR: 0.8809 \\
 & LTE: 43.0111 & LTE: 0.9989 & LTE: 0.0058 & LTE: 0.8389 & SRNO: 44.6216 & CiaoSR: 0.9996 & SRNO: 0.0021 & LTE: 0.8809 \\

\midrule

 & LIIF: 40.7545 & CiaoSR: 0.9988 & LIIF: 0.0143 & CiaoSR: 0.7833 & SRNO: 40.5276 & SRNO: 0.9989 & SRNO: 0.0138 & SRNO: 0.7888 \\
3 & CiaoSR: 40.3483 & LIIF: 0.9987 & CiaoSR: 0.0145 & LIIF: 0.7774 & CiaoSR: 40.5019 & LTE: 0.9989 & LTE: 0.0139 & LTE: 0.7865 \\
 & SRNO: 39.436 & SRNO: 0.9986 & SRNO: 0.0196 & SRNO: 0.744 & LTE: 40.4781 & CiaoSR: 0.9989 & CiaoSR: 0.014 & CiaoSR: 0.7864 \\

\midrule
 
 & SRNO: 38.0577 & CiaoSR: 0.9931 & CiaoSR: 0.0354 & CiaoSR: 0.6943 & SRNO: 38.0536 & SRNO: 0.9935 & SRNO: 0.0336 & SRNO: 0.7021 \\
4 & LTE: 38.04 & LIIF: 0.9927 & LIIF: 0.0354 & SRNO: 0.6899 & CiaoSR: 37.9505 & LTE: 0.9934 & CiaoSR: 0.034 & CiaoSR: 0.6991 \\
 & LIIF: 38.0388 & SRNO: 0.9927 & SRNO: 0.0362 & LIIF: 0.6889 & LTE: 37.9144 & CiaoSR: 0.9934 & LTE: 0.0341 & LTE: 0.6983 \\

\midrule
 
 & SRNO: 34.4939 & SRNO: 0.9704 & SRNO: 0.078 & SRNO: 0.5384 & SRNO: 34.5707 & SRNO: 0.9718 & SRNO: 0.0766 & SRNO: 0.5458 \\
6 & LTE: 34.4752 & LTE: 0.9703 & LTE: 0.0784 & LTE: 0.5377 & LTE: 34.4938 & CiaoSR: 0.9716 & CiaoSR: 0.0773 & LTE: 0.5431 \\
 & MetaSR: 34.372 & CiaoSR: 0.9703 & MetaSR: 0.0796 & CiaoSR: 0.5354 & CiaoSR: 34.491 & LTE: 0.9715 & LTE: 0.0777 & CiaoSR: 0.5428 \\

\midrule
 
 & SRNO: 32.1595 & SRNO: 0.9406 & SRNO: 0.1111 & SRNO: 0.4227 & SRNO: 32.2607 & SRNO: 0.9431 & SRNO: 0.1097 & SRNO: 0.4316 \\
8 & LTE: 32.1208 & LTE: 0.9404 & LTE: 0.1117 & LTE: 0.422 & LTE: 32.2203 & CiaoSR: 0.9428 & CiaoSR: 0.1106 & LTE: 0.43 \\
 & CiaoSR: 32.0297 & CiaoSR: 0.9403 & MetaSR: 0.1123 & CiaoSR: 0.4211 & CiaoSR: 32.2005 & LTE: 0.9427 & LTE: 0.1107 & CiaoSR: 0.4294 \\

\midrule

 & SRNO: 29.3461 & SRNO: 0.8758 & MetaSR: 0.1559 & CiaoSR: 0.2846 & SRNO: 29.4948 & SRNO: 0.88 & SRNO: 0.1538 & SRNO: 0.2924 \\
12 & CiaoSR: 29.3355 & CiaoSR: 0.8758 & SRNO: 0.1562 & SRNO: 0.2841 & CiaoSR: 29.4758 & LTE: 0.8796 & LTE: 0.1543 & LTE: 0.2918 \\
 & LTE: 29.3089 & LTE: 0.8754 & LTE: 0.1566 & LTE: 0.2836 & LTE: 29.475 & CiaoSR: 0.8795 & CiaoSR: 0.1545 & CiaoSR: 0.2915 \\

\bottomrule
\end{tabular}
\caption{Summary of results achieved on scene text domain-specific SVT dataset.}
\label{tab:domainres2}
\end{table*}

\subsection{Bias in IQA reporting: Y-Channel vs. RGB evaluation}
\label{sec:evalbias}
In the SISR domain, it is common practice to evaluate PSNR, SSIM, and other IQA metrics on the luminance (Y) channel of the YCbCr representation, rather than directly on the generated RGB image~\cite{li2023ntire, yoo2020rethinking}. The supporting rationale behind this is that human perception is more sensitive to luminance variations than to color, and the Cb and Cr channels primarily encode color information, which has less impact on capturing high-frequency details in the super-resolved image.

\begin{figure*}[!htb]
  \centering
   \includegraphics[width=0.97\linewidth]{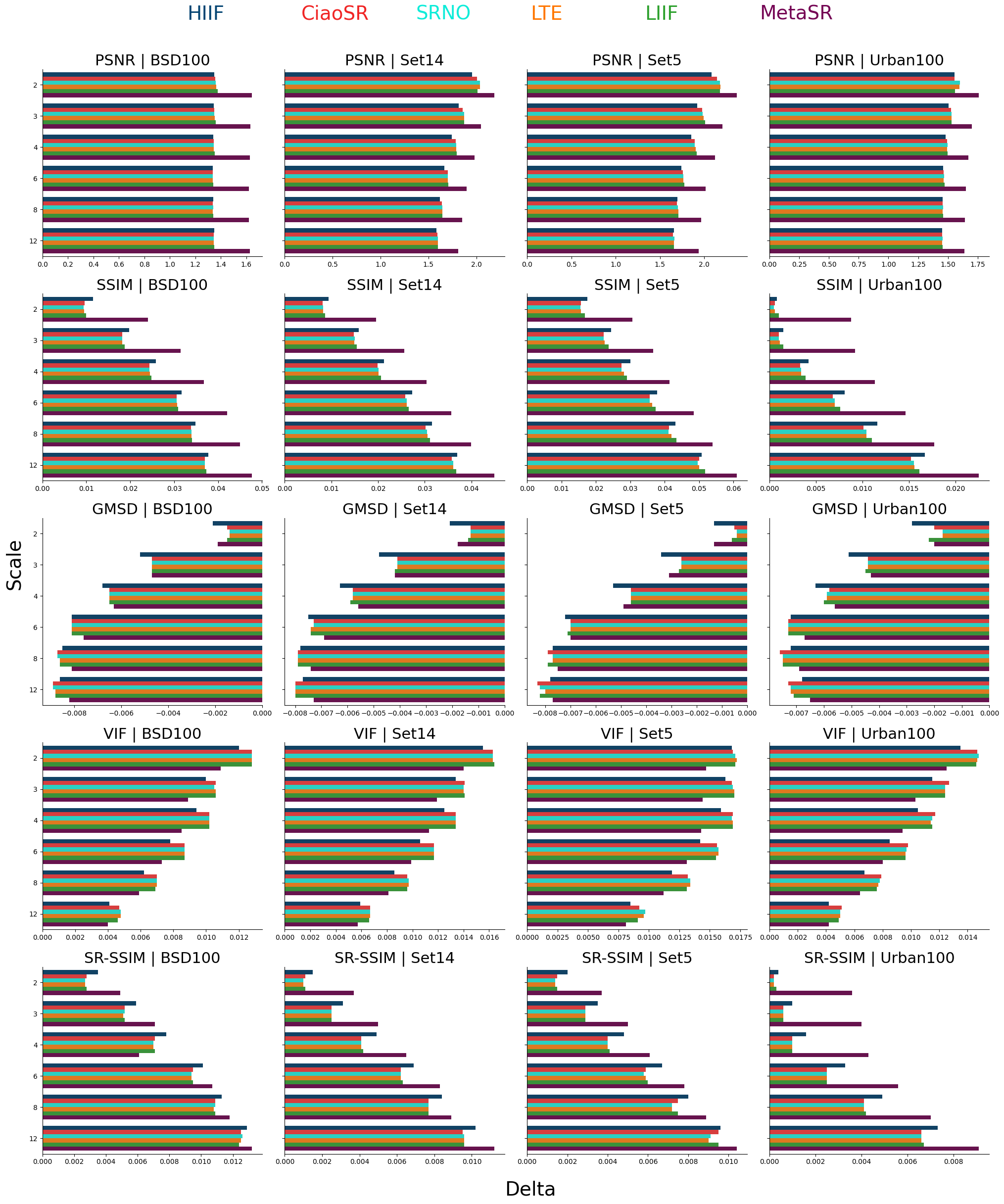}

   \caption{Relative change in IQA when evaluated using the Y-Channel alone for benchmark datasets. EDSR Encoder.}
   \label{fig:edsrrgbychannel}
\end{figure*}

\begin{figure*}[!htb]
  \centering
   \includegraphics[width=0.97\linewidth]{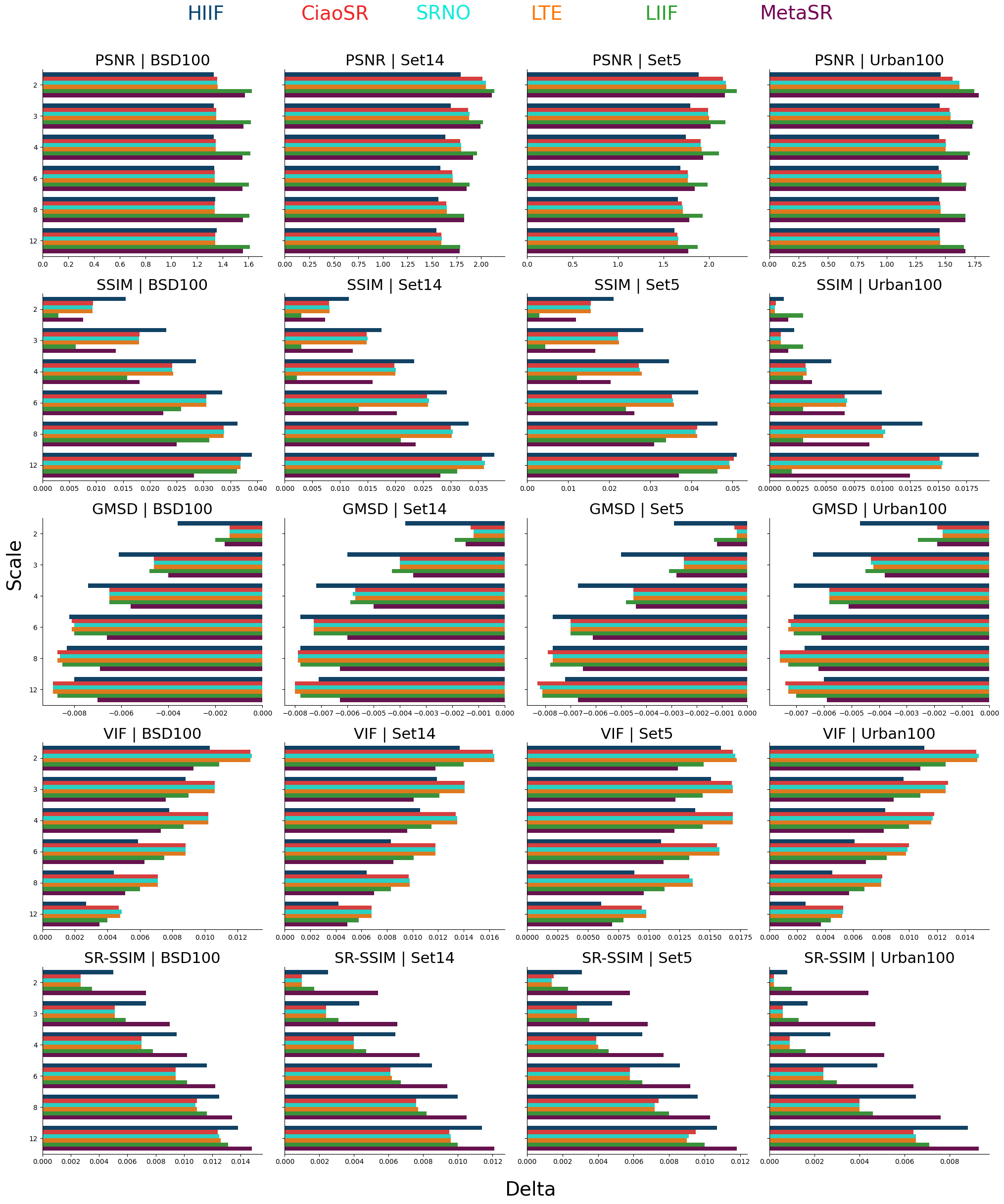}

   \caption{Relative change in IQA when evaluated using the Y-Channel alone for benchmark datasets. RDN Encoder.}
   \label{fig:rdnrgbychannel}
\end{figure*}

Prior works report PSNR for benchmark datasets, Set5, Set14, BSD100, and Urban100, using only the Y-channel, while DIV2K results are typically reported on RGB images~\cite{jiang2025hiif, hu2019meta, lee2022local}. We followed the same convention in all the results presented before. However, to further investigate the impact of this evaluation choice, an additional analysis was conducted comparing IQA metrics computed on the full RGB image versus the Y-channel alone. For each model trained under various settings, results were averaged separately for RGB- and Y-channel-based evaluations across different IQA metrics, scales, and benchmark datasets. Figures~\ref{fig:edsrrgbychannel} and ~\ref{fig:rdnrgbychannel} report the performance gains observed when computing metrics on the Y-channel alone compared to the full RGB image.

The analysis reveals a consistent increase in reported performance when using the Y-channel alone compared to the full RGB image. This highlights an important source of bias in evaluation that, while commonly applied in prior works, is rarely discussed in detail. Note that LPIPS and FSIM are excluded from this comparison, as they are always computed on RGB images and cannot be meaningfully applied to the Y-channel alone.

